\documentclass[AMA,STIX1COL]{WileyNJD-v2}
\usepackage{moreverb}
\usepackage{graphicx}
\usepackage{amsmath}
\usepackage{diagbox}
\usepackage{caption}
\usepackage{subfigure}
\usepackage{algpseudocode}
\usepackage{algorithmicx,algorithm}
\usepackage{caption}
\usepackage{amssymb}
\usepackage{threeparttable}
\usepackage{multirow,booktabs}

\newcommand\BibTeX{{\rmfamily B\kern-.05em \textsc{i\kern-.025em b}\kern-.08em
T\kern-.1667em\lower.7ex\hbox{E}\kern-.125emX}}

\articletype{Research Article}%

\received{2023}
\revised{}
\accepted{}

\begin{document}

\title{UIILD: A Unified Interpretable Intelligent Learning Diagnosis Framework for Intelligent Tutoring Systems}

\author[1]{Zhifeng Wang*}

\author[1]{Wenxing Yan}

\author[2]{Chunyan Zeng*}

\author[1]{Yuan Tian}

\author[1]{Shi Dong}

\authormark{ZHIFENG WANG \textsc{et al}}

\address[1]{\orgdiv{School of Educational Information Technology, Faculty of Artificial Intelligence in Education}, \orgname{Central China Normal University}, \orgaddress{\state{Wuhan 430079}, \country{China}}}

\address[2]{\orgdiv{Hubei Key Laboratory for High-efficiency Utilization of Solar Energy and Operation Control of Energy Storage System}, \orgname{Hubei University of Technology}, \orgaddress{\state{Wuhan 430068}, \country{China}}}

\corres{*Zhifeng Wang, School of Educational Information Technology, Faculty of Artificial Intelligence in Education, Central China Normal University, Wuhan 430079, China. \email{zfwang@ccnu.edu.cn};
\\
*Chunyan Zeng, Hubei Key Laboratory for High-efficiency Utilization of Solar Energy and Operation Control of Energy Storage System, Hubei University of Technology, Wuhan 430068, China. \email{cyzeng@hbut.edu.cn}}


\abstract[Abstract]{Intelligent learning diagnosis is a critical engine of intelligent tutoring systems, which aims to estimate learners' current knowledge mastery status and predict their future learning performance. The significant challenge with traditional learning diagnosis methods is the inability to balance diagnostic accuracy and interpretability. Although the existing psychometric-based learning diagnosis methods provide some domain interpretation through cognitive parameters, they have insufficient modeling capability with a shallow structure for large-scale learning data. While the deep learning-based learning diagnosis methods have improved the accuracy of learning performance prediction, their inherent black-box properties lead to a lack of interpretability, making their results untrustworthy for educational applications. To settle the above problem, the proposed unified interpretable intelligent learning diagnosis (UIILD) framework, which benefits from the powerful representation learning ability of deep learning and the interpretability of psychometrics, achieves a better performance of learning prediction and provides interpretability from three aspects: cognitive parameters, learner-resource response network, and weights of self-attention mechanism. Within the proposed framework, this paper presents a two-channel learning diagnosis mechanism LDM-ID as well as a three-channel learning diagnosis mechanism LDM-HMI. Experiments on two real-world datasets and a simulation dataset show that our method has higher accuracy in predicting learners' performances compared with the state-of-the-art models, and can provide valuable educational interpretability for applications such as precise learning resource recommendation and personalized learning tutoring in intelligent tutoring systems.}



\maketitle

\footnotetext{\textbf{Abbreviations:} DKT, deep knowledge tracing; DKVMN, dynamic key-value memory networks; DINA, deterministic input, noisy ``and'' gate; Ho-DINA, high-order DINA; IRT, item response theory; MIRT, multi-dimensional IRT; LDM-ID, two-channel learning diagnosis mechanism based on the fusion of IRT and DINA; LDM-HMI, three-channel learning diagnosis mechanism based on the fusion of Ho-DINA, MIRT, and IRT; DeepMFLD, deep matrix factorization-based learning diagnosis; NeuralCDM, neural cognitive diagnosis model; IRR, item response ranking; PMF-LDM, probabilistic matrix factorization-based learning diagnosis model; SAE, stacked autoencoder.}

\section{Introduction}
The integration of modern information technology with traditional teaching and education methods has promoted the emergence of smart education, which refers to a new form of education that relies on modern information technology in the education field \cite{hoel_standards_2018}. It applies new technologies such as artificial intelligence, virtual reality, cloud computing, and big data to improve the quality and effectiveness of education \cite{zhu_research_2016-1,Wang2023d}. As a vital part of smart education, the intelligent tutoring system can support monitoring and measuring the learning process and digging out the learning patterns of learners \cite{Gan2022}, as well as effectively presenting the results to teachers and learners \cite{singh2020smart}. Correspondingly, intelligent learning diagnosis \cite{ChenMininglearnerprofile2007} is one of the key components of the intelligent tutoring system \cite{Chuenhancedlearningdiagnosis2010}, and it supports the intelligent tutoring system through data mining and educational theories technically. To provide learners with personalized learning tutoring services, the intelligent learning diagnosis employs educational psychology and data mining technologies to diagnose the learner's learning state and predict possible future learning performances by collecting and analyzing the exercise records and behavioral data, which are generated by the learner during the learning process. In both online and offline education scenarios, the number of teachers is far lower than the number of learners. There is an urgent need for intelligent means to assist teachers in improving efficiency and provide targeted guidance for learners to bridge this resource mismatch. As a result, intelligent learning diagnosis has been applied in many e-learning systems such as Coursera, EDX, Khan Academy, and offline classes, and it has brought great progress to the education industry \cite{burgos2018data}. While facilitating the renewal of teachers' teaching philosophy and driving smart learning systems to provide large-scale personalized learning services, the intelligent learning diagnosis can help teachers provide appropriate guidance based on learners' characteristics, as well as help learners gain a more objective and clear understanding of their learning progress and status, enabling learners to regulate and standardize their learning independently. 

The key to intelligent learning diagnosis is to mine the learner's potential learning state, including their mastery state of knowledge points and learning ability, then predict the learner's performance on particular learning tasks. For this pivotal problem, two types of solutions are widely developed at present: psychometrics-based methods \cite{stout2007skills,hand1998handbook,whitely1980mirt,de2009simultaneous,de2009dina,templin2006measurement,maris1999estimating,de2004hodina} and deep learning-based methods \cite{Piech2015,su2018exercise,shen2020convolutional,zhang2017dynamic,tsutsumi2021deep,nakagawa2019graph,Yang2021c,lu2020towards}. Firstly, psychometrics-based methods appeared earlier and have been developed for more than 40 years. They usually describe the features of the learner and the exercises based on a specific cognitive perspective and employ an empirical response function to express the relationship between learners' performance on exercises and their knowledge status. Psychometrics-based methods are easy to understand, and the pre-defined parameters and the connections between them have clear actual significance and interpretability. Secondly, deep learning diagnosis methods appeared with the development of deep learning technology, which collects the learning resource-related information and learners' learning data, then models learners' learning process and their knowledge status through deep neural networks such as recurrent neural networks \cite{Piech2015}, convolutional neural networks \cite{shen2020convolutional}, memory neural networks \cite{zhang2017dynamic}, graph neural networks \cite{nakagawa2019graph} and so on. With support from deep neural networks' powerful representation learning capabilities, deep learning-based diagnosis methods achieve higher accuracy in learning performance prediction.

The above methods are well developed for intelligent learning diagnosis and have been applied in different scenarios of intelligent tutoring systems. However, there are still some significant challenges. Firstly, traditional psychometric learning diagnosis models introduce a few specific cognitive parameters to model learners from a single perspective, making them difficult to fit the actual learning situation perfectly. For example, the Deterministic Input, Noisy ``And'' gate (DINA) model can only diagnose the learner’s mastery of knowledge in a relatively rough way \cite{de2009dina}. When different learners have not mastered a certain knowledge point, it cannot reflect more detailed differences. Secondly, due to the black box characteristics of deep learning methods, the interpretability of deep learning diagnosis is relatively low. Given an example, the deep knowledge tracing (DKT) model summarizes a learner's state of all knowledge concepts in one hidden state, so that it is difficult to know exactly how much a student has mastered a certain knowledge concept \cite{Piech2015}.

In this paper, we propose a novel unified interpretable intelligent learning diagnosis framework that can effectively solve the above problems. Firstly, multi-channel psychometric learning diagnosis models are introduced to initialize the learning diagnosis of learners from a multi-cognitive perspective. Secondly, the learner representation network and learning resource representation network are constructed separately using the Stacked Autoencoder (SAE) network. Then the learner-learning resource response network is designed to extract deep learning features. Subsequently, deep learning features are fused with shallow learning features, and a self-attention mechanism is introduced to weigh the fused features. Finally, a learning performance prediction network is designed through convolutional neural networks. While obtaining better learning performance predictions, the proposed framework provides interpretability in three aspects: cognitive parameters, learner-resource response network, and weights of self-attention mechanism. Our main contributions can be summarized as follows:

\begin{itemize} 
	\item Framework: The proposed unified interpretable intelligent learning diagnosis framework, which benefits from the powerful representation learning ability of deep learning and the interpretability of psychometrics, achieves a better performance of learning prediction and provides interpretability from three aspects: cognitive parameters, learner-resource response network, and weights of self-attention mechanism. Further, the proposed framework is a generic learning diagnosis pattern that can incorporate a variety of different psychometric models and deep learning diagnosis methods into this framework.  
	\item Mechanisms: Based on the proposed framework, a 2-channel learning diagnosis mechanism LDM-ID is implemented based on the fusion of IRT and DINA on the one hand, and a 3-channel learning diagnosis mechanism LDM-HMI is implemented based on the fusion of Ho-DINA, MIRT, and IRT on the other hand. The proposed two learning diagnosis mechanisms provide reference examples for developing new learning diagnosis models in different educational scenarios. The codes of these two mechanisms are released at \url{https://github.com/CCNUZFW/LDM-ID-HMI}.
	\item Application: The proposed two mechanisms are compared with 11 state-of-the-art methods in two publicly available online class databases and a self-built offline class database, and the proposed method in this paper obtains better learning prediction accuracy (AUC absolute improves by about 5\%), while the stability of learning prediction is also better (RMSE relatively decreases by about 7\%), and is comparable to current deep learning diagnosis methods in terms of prediction time cost. In addition, this paper constructs and open-sources a database of participants in offline courses named CL21 to verify the effectiveness of this method in both online and offline classes. CL21 is released at \url{https://github.com/CCNUZFW/CL21}.
\end{itemize}

The rest of the paper is organized as follows: Section \ref{RW} briefly reviews some related work on intelligent learning diagnosis. Section \ref{Define} presents the intelligent learning diagnosis model, including the basic assumptions, definitions, and problem formula. Section \ref{framework} introduces our proposed unified interpretable intelligent learning diagnosis framework in detail, and Section \ref{method} describes two specific implementation methods under the proposed framework. Then experiments are conducted on two real-world datasets and one virtual dataset, and the results of the proposed method are compared with several baseline methods in Section \ref{experiment}. Finally, we conclude the paper and identify future research that could be carried out based on our new findings in Section \ref{conclusion}.

\section{Related work} \label{RW}

This section briefly reviews the main research work related to intelligent learning diagnosis, which can be divided into two categories: \textit{1) Learning diagnosis based on psychometrics} and \textit{2) Learning diagnosis based on deep learning.} While the psychometrics-based learning diagnosis methods aim at mining learners' potential cognitive situation, the deep learning-based learning diagnosis methods focus on the deep representation learning of learning data and the accurate prediction of learning performance.

\subsection{Learning diagnosis based on psychometrics}
This category of learning diagnosis methods is derived from psychometrics \cite{furrPsychometricsIntroduction2021}. These methods are based on cognitive psychological theories, using modern statistical methods to construct learning diagnosis models and combining estimation and inference to diagnose the cognitive structure and state of learners \cite{paulsenExaminingCognitiveDiagnostic2021}. Psychometrics-based learning diagnosis methods can be divided into two subcategories: \textit{1) continuous learning diagnosis models} and \textit{ 2) discrete learning diagnosis models}.

\textit{1) Continuous learning diagnosis models}, which are based on Item Response Theory (IRT) \cite{hambletonItemResponseTheory2013}, assume learners have a "latent trait" and learners' responses to exercises are determined by the exercise's factors and the learner's potential features jointly. The learner's knowledge state diagnosis result is demonstrated by modeling the relationship among the learner's potential features, the exercise's factor, and the learner's answering response \cite{stout2007skills}. Typical methods, such as IRT \cite{hambletonItemResponseTheory2013} use logical functions to describe the relationship between learner parameters and responses, but the learner characteristic parameters in this model are an expression of the learner's overall mastery level and cannot be refined to every knowledge point. As a result, a Multi-dimensional IRT (MIRT) model \cite{whitely1980mirt} was developed, which estimates the ability of learners in multiple dimensions at the same time, and considers the relationship between each ability dimension. Furthermore, since MIRT does not consider the hierarchical relationship between multi-dimensional ability traits, de la Torre et al. proposed a high-order item response model (Ho-IRT) \cite{de2009simultaneous}, which can directly deal with the potential traits of learners with a hierarchical structure. 

\textit{2) Discrete learning diagnosis models}, which assume that the ability value is discontinuous, and the latent knowledge space is composed of $ K $ dichotomous variables. Therefore, there are $ 2^K $ knowledge mastery states. The learners are divided into these mastery states according to their exercise responses and distinguish their cognitive status of knowledge accordingly. For example, the DINA \cite{de2009dina} model uses guess and slip parameters to define the attributes of the exercise, and it employs a knowledge mastery mode vector that takes the value of 0 or 1 to represent the learner's mastery of each knowledge point. It is assumed that the learner must master all the knowledge points corresponding to the exercise to answer correctly based on the non-compensation assumption. Correspondingly, based on compensation assumption, the Deterministic Input, Noisy ``Or'' gate (DINO) \cite{templin2006measurement} model assumes that the learner can make the correct answer after mastering at least one of the knowledge points. Further, for more fine-grained modeling, the Noisy Input, Deterministic ``And'' gate (NIDA) \cite{maris1999estimating} model defines the parameters at the knowledge points level of the exercises, and refines the scope of the parameters compared with DINA and DINO models. To be able to report both the macroscopic general ability and the microscopic cognitive state at the same time, the High-order DINA (Ho-DINA) model \cite{de2004hodina} is proposed, which assumes that the knowledge points are independent of each other and subordinate to higher-order general capabilities. Assuming that the learner's response matrix is determined by the inner product of the learner's knowledge proficiency matrix and the knowledge point vector contained in the exercise, the Probabilistic Matrix Factorization-based Learning Diagnosis Model (PMF-LDM) is established through matrix decomposition \cite{Mnih2007}.

Although psychometric-based learning diagnostic methods can provide explanations of learning diagnosis in terms of cognitive parameters, current methods define targeted cognitive parameters for a specific learning scenario. These learning diagnostic models based on specific cognitive parameters can only interpret learning diagnostic results from a single perspective, so it becomes important to construct a unified learning diagnosis framework that includes multiple cognitive parameters from integrated perspectives and can be applied to multiple learning scenarios.

\subsection{Learning diagnosis based on deep learning}
Deep learning methods have powerful representation capabilities and have been widely used in computer vision \cite{Zeng2023c,Wang2022ac,Zeng2022,Wang2023a,Li2023,Wang2022at,Zeng2021c,Wang2021,Zeng2020a,Tian2018a}, speech processing \cite{Zeng2023b,Wang2022t,Zeng2021a,Wang2021m,Zeng2023,Wang2020h,Zeng2022a,Wang2018a,Zeng2021b,Wang2015b,Zeng2020,Zhu2013,Zeng2018,Wang2011,Wang2011a,Zeng2023b,Zeng2023}, natural language understanding \cite{floridiGPT3ItsNature2020}, data mining \cite{Wang2023,Li2023a,Lyu2022}, and other fields \cite{Wang2017,Wang2022as,Min2019,Min2018,Wang2015a}. 

Currently, advanced deep learning methods have been widely studied in diagnosis and prediction tasks. For intelligent fault diagnosis, the advent of deep learning has constructed an end-to-end diagnosis procedure \cite{Lei2020}. Further, Out-of-distribution detection-assisted trustworthy machinery fault diagnosis approach is developed to enhance the reliability and safety of intelligent diagnosis models \cite{Han2022d}. For the wheel wear status diagnosis of high-speed trains, a multiplex local temporal fusion architecture is combined with the transformer architecture \cite{Wang2022aq}.

Deep learning is also extensively used in the learning diagnosis task of learners in the field of intelligent tutoring systems \cite{Lyu2022}. Deep learning-based learning diagnosis usually models the learner's learning process based on their answering records sequence through an end-to-end trainable neural network. Classified by the network structures, the learning diagnosis methods based on deep learning could be divided into five categories: \textit{1) The recurrent neural network-based learning diagnosis}, \textit{2) The convolutional neural network-based learning diagnosis}, \textit{3) The memory network-based learning diagnosis}, \textit{4) The graph neural network-based learning diagnosis.}, and \textit{5) The deep neural network-based learning diagnosis.}

\textit{1) The recurrent neural network-based learning diagnosis:} The DKT proposed by Piech et al. \cite{Piech2015} first applied the recurrent neural network to learning diagnosis, which could break the restriction of empirical response function by simulating the interaction between the learner and the exercise with recurrent neural networks. However, DKT can only simply predict whether a learner has mastered or not mastered a specific knowledge point, and can do nothing about the intermediate state of learner knowledge mastery. Further, Su et al. \cite{su2018exercise} proposed the Exercise-Enhanced Recurrent Neural Network (EERNN) that predicts the learner’s performance through the learner’s answer record and the text information of each exercise. For multi-features, Xiong et al. \cite{2016Going} proposed an extended DKT model, adding more answering features such as answering time, question difficulty, learners' previous knowledge, and so on to assist in predicting learning performance. In terms of the improvement of the model structure, Shalini et al. \cite{2019A} proposed the Self-attentive Knowledge Tracing (SAKT) model, which is the first to apply the attention mechanism to the learning diagnosis task. Besides, Aritra et al. \cite{2020Context} used a monotonic attention mechanism to limit the weight of questions that are far away from the learner's current answer. Based on the attention mechanism, Youngduck et al. \cite{2020Towards} further applied the encoder-decoder model to the learning diagnosis task.

\textit{2) The convolutional neural network-based learning diagnosis:} Different from models based on recurrent neural networks, to achieve personalized modeling, Shen et al. \cite{shen2020convolutional} proposed the convolutional knowledge tracing (CKT) method, which utilizes the sensitivity of convolutional neural networks for the temporal and spatial information to consider the prior knowledge and learning rate for each learner. Further, Wang et al. \cite{2020Convolutional} combined a convolutional neural network and a recurrent neural network into an integrated model to build Convolutional Recurrent Knowledge Tracing (CRKT).

\textit{3) The memory network-based learning diagnosis:} Since DKT only uses a latent state to represent the learner’s knowledge mastery state and cannot simulate the dynamic change process of learners' mastery of different knowledge points, Zhang et al. \cite{zhang2017dynamic} proposed Dynamic Key-Value Memory Networks (DKVMN), which draws on the idea of memory networks, uses a key matrix for representing the knowledge and a value matrix for indicating the learner’s mastery of each knowledge point and predicts learner's performance according to the key matrix and value matrix. Furthermore, Tsutsumi et al. \cite{tsutsumi2021deep} add exercises’ difficulty and learners’ ability to DKVMN to improve the accuracy and interpretability.

\textit{4) The graph neural network-based learning diagnosis:} Inspired by cognitive theory, the structural information between knowledge points is also helpful for learning diagnosis. Therefore, Nakagawa et al. \cite{nakagawa2019graph} explored the knowledge graph structure and applied it to learning diagnosis with graph neural networks. Besides, Yong Yu et al. \cite{Yang2021c} considered the exercise-knowledge correlations for learning diagnosis through graph convolutional networks. At the same time, Tong et al. \cite{2020Structure} are more focused on using graph neural networks to capture the relationship between different knowledge points, such as similarity or prerequisite relationship. Further, Abdelrahman et al. \cite{2021Deep} used the latent knowledge graph to improve the reading and writing process of the memory matrix of the DKVMN model.

\textit{5) The deep neural network-based learning diagnosis:} Neural Cognitive Diagnosis Model (NeuralCDM) \cite{Wang2022an} is proposed to project students and exercises to factor vectors and incorporates neural networks to obtain complex learning interactions. Besides, Deep Matrix Factorization-based Learning Diagnosis (DeepMFLD) \cite{Xue2017} acquires the learners' embedding and the exercises' embedding separately by deep matrix decomposition, then fuses them with a deep neural network, and finally predicts the learner's performance by a fully connected layer. Further, Item Response Ranking (IRR) framework \cite{Tong2021a} introduces pairwise learning into learning diagnosis to sufficiently model the monotonicity between learning interactions.

Although the accuracy of learning diagnosis can be improved by introducing deep learning technology, its black-box and end-to-end properties in learning modeling \cite{lu2020towards} lead to its inability to provide supporting explanatory information and model intermediate feedback for instructional and learning decisions.

\subsection{Research questions and motivations}
Based on the research of related work, we summarize the research questions and motivation of this paper as follows:
\begin{itemize} 
	\item
	RQ1: How to build a unified learning diagnosis framework that incorporates multi-dimensional cognitive parameters and can be applied to a broader range of learning scenarios?
	
	Inspired by the multimodal approach, we fuse multiple psychometrics-based learning diagnosis models containing different cognitive parameters into a unified multimodal learning diagnosis model. On the one hand, this allows for multidimensional interpretable cognitive parameters. On the other hand, it also enables the proposed unified learning diagnosis framework to be extended to a more practical and broader range of educational application scenarios.
	
	\item 
	RQ2: How to effectively characterize learners and learning resources and appropriately establish effective interactive responses between learner and learning resource features?
	
	Inspired by the representational power of deep learning methods, on the one hand, we construct a learner network and a learning resource network based on shallow learning features to extract better representational deep learner features and deep learning resource features; on the other hand, we characterize the interactive responses between them by constructing a learner-learning resource response network. Further, the prediction network is constructed by combining shallow and deep learning features to obtain better representativeness as well as to provide interpretable cognitive parameters.
	
	\item 
	RQ3: How to effectively improve learning performance predictions while also providing appropriate explanatory information that can be used to improve the acceptability and usability of learning diagnosis?
	
	Inspired by the interpretable AI approach, first, we provide self-contained explanatory information describing learners' cognitive states from multidimensional cognitive parameters; second, we provide explanatory information on the interaction responses between learners and learning resources by constructing learner-learning resource response networks for discovering which learning interaction channels are more effective; and finally, we obtain global explanatory information on the importance of different learning features in terms of decision making due to the attention mechanism for identifying essential learning cues.
	
\end{itemize}

\section{Intelligent learning diagnosis model} \label{Define}
This section introduces some assumptions and definitions, and then details the problem definition. 

\subsection{Assumptions and definitions of intelligent learning diagnosis}
First, this paper is based on the following assumptions and definitions:
\\ \hspace*{\fill} \\
\textbf{Assumption 1. }The obtained online and offline learning data is used with permission;
\\ \hspace*{\fill} \\
\textbf{Assumption 2. }The data is authentic and complete. The learners participating in the data collection are required to take the course seriously and use their initiative to complete each exercise, to ensure that there is no halfway exit leading to incomplete data or inconsistency in answering results. And the data can reflect the actual level of learners;
\\ \hspace*{\fill} \\
\textbf{Assumption 3. }The learner's answer results are partially independent. On the one hand, it means that the learner's answering results of course exercises will not be affected by other people. At the same time, it will not affect others. On the other hand, it means that the answering results of learners in different exercises will not be affected by each other. 
\\ \hspace*{\fill} \\
\textbf{Definition 1. }\emph{Shallow Feature of Learning Resource}. For the learning resources involved in the course (mainly refer to the exercises done by learners), it obtains descriptive parameters for them based on different educational interpretation meanings to form a parameter matrix $ EC \in R ^ {e \times d_e} $, where $ e $ and $ d_e $ represent the number of exercises and the dimension of shallow learning resource feature respectively;
\\ \hspace*{\fill} \\
\textbf{Definition 2. }\emph{Shallow Feature of Learners}. Combining multiple focal points of different cognitive diagnosis models, it describes learners from different angles and forms a parameter matrix $ SC \in R^{s \times d_s} $, where $ s $ and $ d_s $ respectively represent the number of learners and the dimension of shallow learner feature; 
\\ \hspace*{\fill} \\
\textbf{Definition 3. }\emph{Deep Feature of Learning Resources and Learners}. Based on the two descriptive parameters, it introduces deep learning technology to mine higher-order bottleneck features, which models and characterizes the response between learning resources and learners more effectively.

\subsection{Problem description}
According to the above definitions and assumptions, the unified interpretable intelligent learning diagnosis task is described as follows. Given that $ S $ learners study the same course and complete the same exercise $ E $ corresponding to course $ C $, the multiple psychometric models are integrated to obtain the learner and learning resource parameter matrix, and the learner's answer performance and cognitive state are predicted based on the parameter matrix. Among them, $ S $ is a set of learners, $ E $ is a set of exercises, $ K $ is a set of knowledge points, and the answer record of learner $ s $ is $ R_s=\{(e_1,r_1),(e_2,r_2),...,(e_t,r_t)\} $. Among them: $ e_t $ is the exercises done by the learner $ s $ at time $t$, $ r_t $ is the corresponding answer result, $ r_t\in\{0,1\} $, $ r_t=1 $ means the answer is correct, $ r_t=0 $ means the answer is wrong. $ q_i $ is the set of knowledge points involved in exercise $ i $, $ q_{ij}=1 $ means that there is a correlation between exercise $ i $ and knowledge point $ j $, otherwise, it is irrelevant. The set of answer records of all learners $ R=\{R_1,R_2,...,R_s\} $ consists the answer matrix $ R $, and $ Q=\{q_1,q_2,...,q_i\} $ consist of the exercise-knowledge point association matrix $ Q $. Then the problem can be defined as:

\textbf{Definition 4. }\emph{Intelligent learning diagnosis problem}. Given the exercising logs of each learner and the knowledge points involved in each exercise, our goal is threefold: (1) extract interpretable parameters of learners and learning resources; (2) construct a learner-resource response network and deep-shallow feature fusion mechanism to obtain the explanatory information on the importance of learning features; (3) predict the learning performance $\widehat{r_{t+1}}$ of the specific learner on time $t+1$ and offer interpretable information for decision-making.

\begin{figure}[!h]
	\centering
	\includegraphics[width=6.7in]{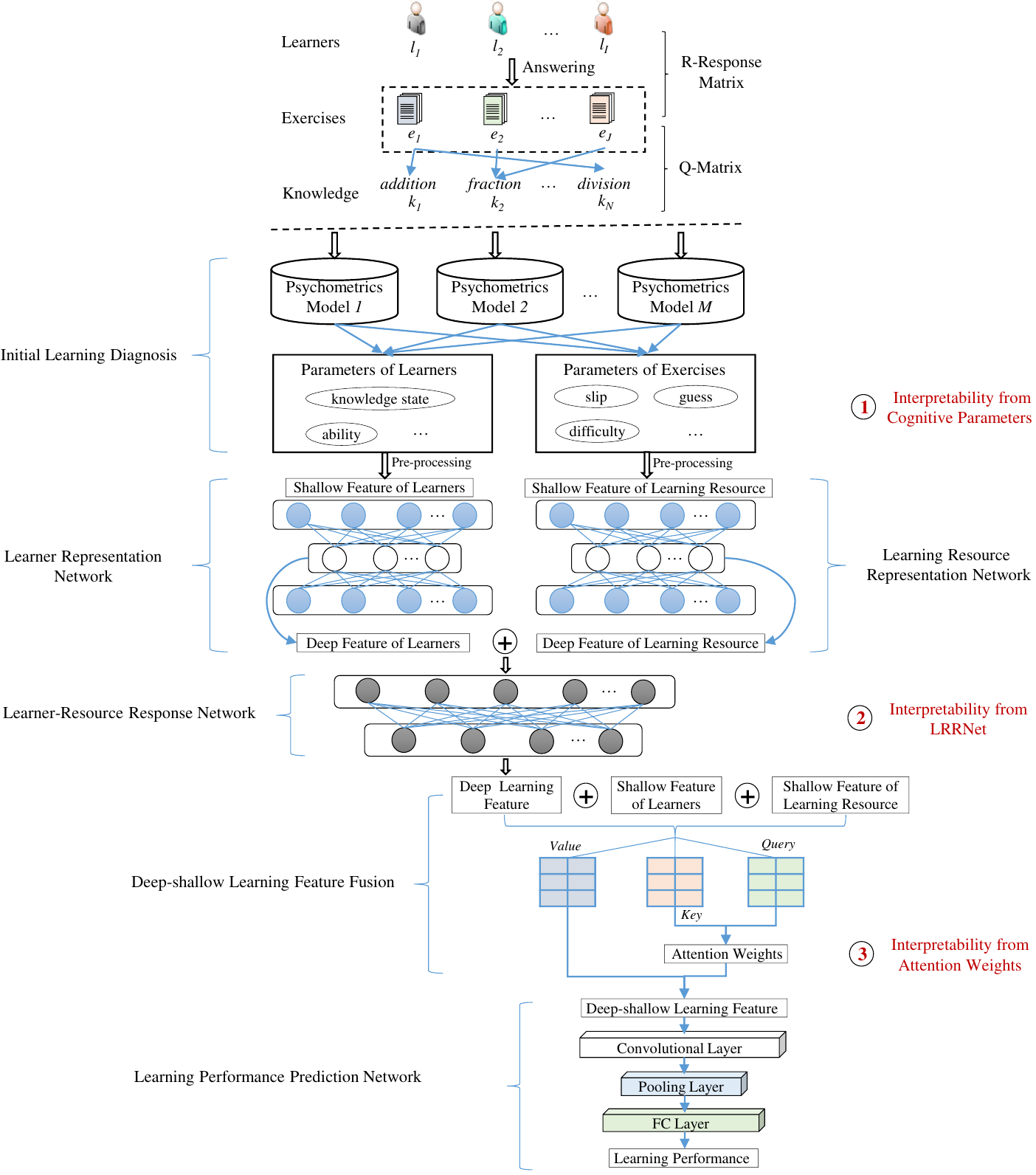}
	\caption{The inputs of the proposed unified interpretable intelligent learning diagnosis framework are the learners' response matrix $R$ and the exercise-knowledge point matrix $Q$. The proposed framework contains five main components: 1) initial learning diagnosis, 2) learner and learning resource representation network, 3) learner-resource response network, 4) deep-shallow learning feature fusion, and 5) learning performance prediction network. The interpretability can be obtained from three aspects: 1) the interpretability from cognitive parameters, 2) the interpretability from LRRNet, and 3) the interpretability from attention weights.}
	\label{fig1}
\end{figure}

\section{Proposed unified interpretable intelligent learning diagnosis framework} \label{framework}
Based on the assumptions and definitions of the above learning diagnosis model, this paper proposes a unified interpretable learning diagnosis framework, which is versatile and can flexibly formulate specific strategies according to different needs to realize the interpretable learning diagnosis and performance prediction of learners. As shown in Figure 1, the proposed framework mainly consists of five components: initial learning diagnosis, learner and learning resources representation network, learner-resource response network, deep-shallow learning feature fusion, and learning performance prediction network. The framework that takes advantage of the powerful representation learning ability of deep learning and the interpretability of psychometrics, achieves a good performance of learning prediction and provides interpretability from three aspects: cognitive parameters, learner-resource response network, and weights of self-attention mechanism.

\subsection{Initial learning diagnosis}
The main purpose of the initial learning diagnosis module is to construct the multi-channel cognitive parameter set.

The cognitive parameter set construction process is shown in Figure 1. Based on the learner's historical learning records and learning resource information, the multi-channel psychometric model is used to make a preliminary diagnosis of the learner, and to estimate the parameters of the learning resource at the same time, thereby constructing the learning resource parameter set $ EC $ and learner parameter set $ SC $. 

Taking into account the interpretability of the learner's cognitive state, it is worth noting that the parameters are highly interpretable with practical significance in cognitive theory and psychometrics, such as the difficulty and discrimination of the exercise, the ability of the learner, which differs from learning parameters through network training in the deep learning-based learning diagnosis methods. 

\begin{gather}
	EC=\{PM_1 (R,Q),PM_2 (R,Q),...,PM_n (R,Q)\}\\
	SC=\{PM_1 (R,Q),PM_2 (R,Q),...,PM_n (R,Q)\}
\end{gather}

$PM_n$ is the $n^{th}$ psychometric model. Then, based on the answer record $ R_s=\{(e_1,r_1),(e_2,r_2),...,(e_t,r_t)\} $ of learner $ s $, obtain the learner parameter vector $ SC_s $ and the learning resource parameter vector $ EC_1 $ to $ EC_t $ related to learner $ s $.


\subsection{Learner and learning resources representation network} 
This module consists of two parts: learners representation network and learning resources representation network.

Specifically, given the  learner parameter vector $ SC_s $ of learner $ s $ and the learning resource parameter vector $ EC_e $ of exercise $ e $ , firstly, in order to eliminate the difference between the ranges of various parameters, the parameter values are discretized and One-Hot encoded, and then $ SC_s $ is coded as a $ d_0 $ dimensional 0-1 vector $ x^s $, and $ EC_e $ is coded as a $ d_1 $ dimensional 0-1 vector $ x^e $.

To better represent learners and learning resources based on cognitive parameters, the learner representation network and learning resource representation network are designed respectively, which take advantage of auto-encoder to excavate the deep learning features of learners and learning resources. The SAE can copy the input to the output through a hidden layer. It consists of two parts: an encoder that can be represented by the function $ h=f(\cdot) $ and a decoder that generates reconstruction information $ y=g(\cdot) $, by training the auto-encoder to reproduce the input, $ h $ can obtain some useful features, to achieve the purpose of feature learning for the learner's parameter vector. The learners and learning resource auto-encoder designed in this paper can be expressed as:

\begin{gather}
	h^s=f(x^s)=\sigma (W_1^s \times x^s+b_1^s)\\
	y^s=g(h^s)=\sigma (W_2^s \times h^s+b_2^s)\\
	h^e=f(x^e)=\sigma (W_1^e \times x^e+b_1^e)\\
	y^e=g(h^e)=\sigma (W_2^e \times h^e+b_2^e)
\end{gather}
$W, b$, and $\sigma (\cdot)$ are weights, bias, and activation function. We take the intermediate results $ h^s $ and $ h^e $ of the coding layer as the feature learning results, which can represent a more valuable part of the information in the original parameter vector.

\subsection{Learner-resource response network}
Then, the learner-resource response network is used to simulate the answer process in the actual situation. It takes the deep characteristics of learners and learning resources as the input, and outputs the deep characteristics after the interaction between learners and learning resources.

\begin{gather}
	f=concat(h^s,h^e)\\
	f_d=\sigma (W_3 \times f+b_3)
\end{gather}

Among them, $\sigma$ is the $ tanh $ activation function, and $ h $ and $ y $ are the encoder and the decoder, respectively. In addition, the learner parameter encoder's weight matrix $ W_1^s \in R^{d_2 \times d_0}$, the bias vector $ b_1^s \in R^{d_2} $, the decoder's weight matrix $ W_2^s \in R^{d_0 \times d_2} $, bias vector $ b_2^s \in R^{d_0}$; the learning resource parameter encoder's weight matrix $ W_1^e \in R^{d_3 \times d_1} $, bias vector $ b_1^e \in R^{d_3} $, decoder's weight matrix $W_2^e \in R^{d_1 \times d_3} $, the bias vector $b_2^e \in R^{d_1} $ are the parameters of the learner representation network and learning resource representation network. The weight matrix $W_3 \in R^{(d_2+d_3) \times d_4} $ , the bias vector $b_3 \in R^{d_2+d_3}$ are the parameters of the learner-resource response network, $ f_d $ is the deep feature of learners and learning resources.

\subsection{Deep-shallow learning feature fusion}
To improve the performance and interpretability of the model, we regard the original cognitive parameter vector as shallow features and the deep characterization features as deep features, and then we combine the two features as the overall features for diagnosis and prediction.
\begin{equation}
	f=concatenate(f_d,SC_s,EC_e) 
\end{equation}

Among them, the dimensions of $ f $ are $ d_5 $, in which $ d_5=d_4+d_e+d_s $.

In terms of interpretability for the prediction process, we focus on the reasons why the model makes such predictions. Therefore the prediction network in this paper employs the self-attention mechanism to process the fusion features of the learner and the learning resource, in order to fully mine the correlation and importance information between the features of each dimension, so as to provide interpretability from the feature level for the final diagnosis and prediction results, in other words, we can know exactly which features have a more significant impact on the final prediction result.

Use a convolutional layer with a convolution kernel size of 1 to obtain the query vector matrix Query, the key vector matrix Key, and the value vector matrix value through $ f $:
\begin{gather}
	Query=Conv(f)\\
	Key=Conv(f)\\
	Value=Conv(f)
\end{gather}
The weight of each dimension of data in the feature is calculated by calculating the dot product between the query vector matrix and the key vector matrix, and the weight indicates the degree of relevance between the task to be queried and each input data.

\begin{equation}
	similarity(Query,Key_i)=Query \bullet Key_i
\end{equation}	
On the one hand, numerical conversion of the degree of relevance can be carried out to normalize and organize the original calculated scores into a probability distribution with the sum of weights being 1. On the other hand, it can also highlight the weights of essential elements.
\begin{equation}
	a_i=SoftMax(similarity_i)=\frac {e^{similarity_i}}{\sum _{j=1}^{d_5} e^{similarity_j }}
\end{equation}	
The input data is weighted and summed to obtain the fused feature data.
\begin{equation}
	f_a=\sum_{i=1}^{d_5} a_i \cdot Value_i
\end{equation}	

\subsection{Learning performance prediction network}
With the features acquired from the feature extraction process, we construct a learner performance prediction network based on a convolutional neural network, use learner feature vectors and learning resource feature vectors to predict the probability of learners answering a specific exercise correctly, diagnose the knowledge mastery of learners and analyze the learning resources through the feature information of learners and learning resources. 

Use the convolutional layer to extract the spatial information of the deep-shallow fusion feature, and set the convolution kernel size to 3, and the step size to 1. Then add the \textit{Relu} activation function to reduce the dependence between parameters, avoid overfitting, and use the maximum pooling layer to reduce the feature dimension, then obtain the predicted value $p$ of the learner's answer under a specific learning task accordingly. It can be expressed as:
\begin{gather}
	f_c=Conv(f_a)\\
	f_{re} =Relu(f_c)\\
	f_p=MaxPool(f_{re})\\
	p=\sigma(W_4 \times f_p+b_4)
\end{gather} 
$f_c$, $f_{re}$, and $f_p$ are the intermediate variables of the prediction network. To learn the various parameters of the model, use the cross-entropy loss function between the student's real answer value $r_t$ and the predicted answer value $p_t$ as the objective function $L$ of the model:
\begin{equation}
	L=-\sum_{i=1}^{N} r_tlogp_t+(1-r_t)log(1-p_t)
\end{equation}

\subsection{Interpretability of the proposed framework}
The interpretability of our proposed framework includes three aspects, which correspond to the three modules of the framework:
\begin{itemize} 
	\item \textit{Interpretability from cognitive parameters}: Based on the learner's historical learning records and learning resource information, the multi-channel psychometric model is used to make a preliminary diagnosis of the learner, and to estimate the parameters of the learning resource at the same time, thereby constructing the learning resource parameter set $EC$ and learner parameter set $SC$. Taking into account the interpretability of the learner's cognitive state, it is worth noting that the parameters are highly interpretable with practical significance in cognitive theory and psychometrics, such as the difficulty and discrimination of the exercise, the ability of the learner, which differs from learning parameters through network training in the deep learning-based learning diagnosis methods. 
	\item \textit{Interpretability from learner-resource response networks}: In this module, we first characterize the learners and learning resources, and then construct the learner-resource response network. The learner-resource response network is used to simulate the complex learning activities in the process of learners' answers, that is, the interactive response between learners and learning resources, which is closer to the actual situation and has a better interpretability of the learning process. 
	\item \textit{Interpretability from weights of self-attention mechanism}: In terms of interpretability for the prediction process, we focus on the reasons why the model makes such predictions, therefore the prediction network in this paper employs the self-attention mechanism to process the fusion features of the learner and the learning resource, in order to fully mine the correlation and importance information between the features of each dimension, so as to provide interpretability from the feature level for the final diagnosis and prediction results, in other words, we can know exactly which features have a greater impact on the final prediction result.
\end{itemize}

\section{Multi-channel intelligent learning diagnosis mechanisms} \label{method}
\begin{figure}[!h]
	\centering
	\includegraphics[width=\linewidth]{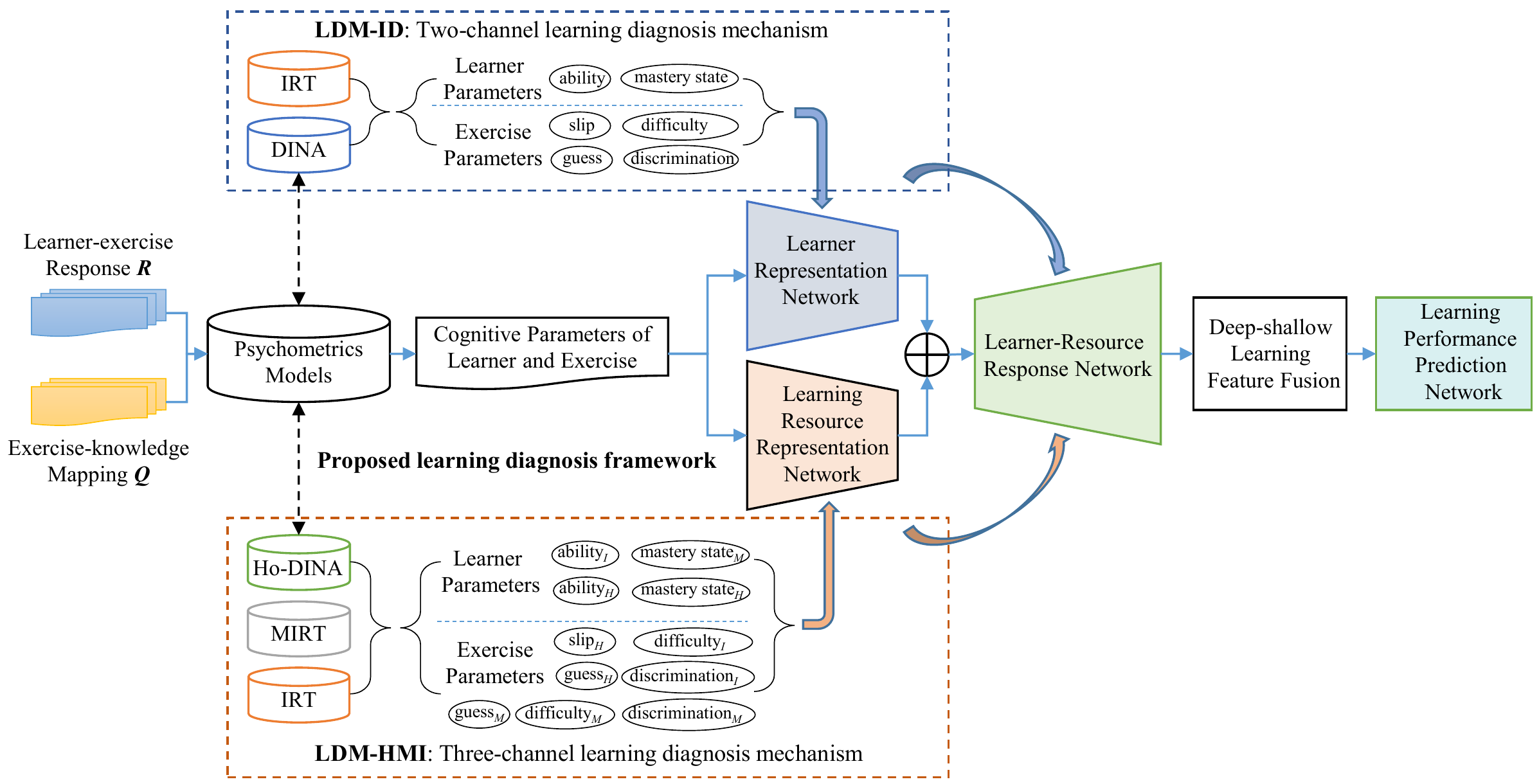}
	\caption{Based on the proposed unified interpretable learning diagnosis framework, two learning diagnosis mechanisms were developed. LDM-ID was based on IRT and DINA, which extracted two categories of learner cognitive parameters and four categories of exercise cognitive parameters. LDM-HMI was based on Ho-DINA, MIRT, and IRT, which extracted four categories of learner cognitive parameters and seven categories of exercise cognitive parameters, respectively.}
	\label{fig1_2}
\end{figure}

Within the proposed unified interpretable intelligent learning diagnosis framework in Section \ref{framework}, two specific multi-channel intelligent learning diagnosis mechanisms are implemented in this section. As shown in Figure \ref{fig1_2}, the first implementation is a two-channel learning diagnosis mechanism based on the fusion of IRT and DINA (LDM-ID), and the other is a three-channel learning diagnosis mechanism based on the fusion of Ho-DINA, MIRT, and IRT (LDM-HMI).

\subsection{Two-channel learning diagnosis mechanism based on the fusion of IRT and DINA (LDM-ID)}
Generally, LDM-ID selects the IRT and DINA models for initial learning diagnosis, DINA model uses a vector to represent the master of the learner on each knowledge, the dimension of the vector is equal to the number of knowledge, with the characteristics of the exercises with the slip rate and guess rate. The IRT model uses only one value to represent the overall ability of the learner and describes the exercises with difficulty and discrimination. The information on the parameters used to describe the learners and learning resources in these two models is integrated to obtain the original input parameter sets $ EC $ and $ SC $, and then it performs subsequent learning diagnostics.

\emph{\textbf{Initial diagnosis models of LDM-ID:}} the process of learning diagnosis from the initial parameter set has been described in detail in Section 4.

The response function of the IRT model is:
\begin{equation}
	P(s_{I\_\theta})=e_{I\_gues}+\frac{1-e_{I\_gues}}{1+e^{-De_{I\_disc}(s_{I\_\theta}-e_{I\_diff})}}
\end{equation}
Among them, $ s_{I\_\theta} $ is the learner's learning ability parameter. The parameter $ e_{I\_guess} $ is generally called the "guessing parameter", that is, when the learner's ability value is very low (for example, close to negative infinity), it is still possible to give the right answer to this problem. $ e_{I\_diff} $ is the problem difficulty parameter, and $ e_{I\_disc} $ is the problem discrimination parameter.

The response function of the DINA model is:
\begin{equation}
	P(s_{D\_\alpha})=e_{D\_gues}^{(1-\eta)}\cdot{(1-e_{D\_slip})}^{\eta}
\end{equation}
Among them, $ s_{D\_\alpha} $ is the learner's mastery vector for each piece of knowledge, $ \eta $ is the learner's ideal answer situation under the corresponding mastery state, $ e_{D\_slip} $ and $ e_{D\_gues} $ are the guess and slip parameters of the exercises respectively, and the slip parameter indicates that the learner has mastered all the knowledge required by the exercises he answered, but the probability of answering incorrectly for some reason. The guessing parameter is the probability that the learner has not mastered all the knowledge needed to answer the exercises or even not mastered one knowledge, but answered correctly. The specific parameter estimation process is shown in steps 1-7 of algorithm 1.

\begin{algorithm}[!h]
	\caption{Two-channel learning diagnosis mechanism based on the fusion of IRT and DINA (LDM-ID)} 
	\hspace*{0.02in} {\bf Input:} 
	Student exercise response matrix $R$ and question-knowledge corelation matrix $Q$\\
	\hspace*{0.02in} {\bf Output:}
	Student exercise score prediction and interpretable parameters
	\begin{algorithmic}[1]
		\State 	Initialize parameters matrix $EC$ and $SC$:
		\State $EC^{(0)}=\{e_{I\_diff}^{(0)},e_{I\_disc}^{(0)},e_{D\_gues}^{(0)},e_{D\_slip}^{(0)}\} $, $SC^{(0)}=\{s_{I\_\theta}^{(0)},s_{D\_\alpha}^{(0)}\}$
		
		\While{not converge or not reach iterations}
		\State $ L(R)=\prod_{u=1}^U L(R_u)=\prod_{u=1}^U \sum_{l=1}^L P(R_u|\alpha_l)P(\alpha_l) $
		\State$ L(R|\delta,\pi)=\prod_{k=1}^N\prod_{j=1}^M p(\pi_k|\delta_j)^{r_{jk}}[1-p(\pi_k|\delta_j)]^{1-r_{jk}}$\\ 
		
		// update parameters in DINA
		\State$ e_{D\_slip}=\hat{e}_{D\_slip},e_{D\_gues}=\hat{e}_{D\_gues}\leftarrow \frac{\partial \log L(R)}{\partial e_{D\_slip}}=0,\frac{\partial \log L(R)}{\partial e_{D\_gues}}=0$ \\
		
		// update parameters in IRT		
		\State$ e_{I\_disc}=\hat{e}_{I\_disc},e_{I\_diff}=\hat{e}_{I\_diff},s_{I\_\theta}=\hat{s}_{I\_\theta}\leftarrow \frac{\partial \log L(R)}{\partial e_{I\_disc}}=0,\frac{\partial \log L(R)}{\partial e_{I\_diff}}=0,\frac{\partial \log L(R)}{\partial s_{I\_\theta}}=0 $ 
		
		\EndWhile
		
		\For{student $s$ and exercise $e$}\\
		// extract student features and exercise features via auto-encoder 
		\State $h^s=f(SC_s)=\sigma(W_1 ^s\times SC_s+b_1^s)$
		\State $y^s=g(h^s)=\sigma(W_2 ^s\times h^s+b_2^s)$
		\State $h^e=f(EC_e)=\sigma(W_1 ^e\times EC_e+b_1^e)$
		\State $y^e=g(h^e)=\sigma(W_2 ^e\times h^e+b_2^e)$
		
		\State $h^s(\cdot),y^s(\cdot)=\mathop{\arg\min}\limits_{h^s(\cdot),y^s(\cdot)}||h^s-SC_s||^2,h^e(\cdot),y^e(\cdot)=\mathop{\arg\min}\limits_{h^e(\cdot),y^e(\cdot)}||h^e-EC_e||^2 $
		
		\State $ ((SC_s,h^s),(EC_e,h^e))\rightarrow(exercise,student),f=concat(exercise,student) $\\
		//calculate keys queries and values in attention function 
		\For{$l$ in $len(f)$}
		\State $ Query = Conv(f) Key = Conv(f) Value=Conv(f) $  
		\State $ f_a=\sum_{i=1}^{d_5} \frac{e^{similarity_i}}{\sum_{j=1}^{d_5}e^{similarity_j}}\cdot Value_i$
		\State $ similarity(Query,Key_i)=Query\bullet Key_i $
		\EndFor
		\EndFor
		\State $ f_p=MaxPool(Relu(Conv(f_a))) $
		\State $ p=\sigma(W_3\times f_p+b_3) $
		\State \Return $p$, $EC$, $SC$, attention weights
	\end{algorithmic}
	\label{algorithm1}
\end{algorithm}

Accordingly, a parameter set of learners and a parameter set of learning resources are constructed based on the learner's learning record data and learning resource data:
\begin{gather}
	EC=\{e_{I\_diff},e_{I\_disc},e_{D\_gues},e_{D\_slip}\}\\
	SC=\{s_{I\_\theta},s_{D\_\alpha}\}
\end{gather} 

\emph{\textbf{Representation and prediction network of LDM-ID:}} we model learners and learning resources separately and construct a performance prediction network to complete the learning diagnosis task as described in Section 4 of the paper. The process is described in steps 12-28 of Algorithm 1.

\subsection{Three-channel learning diagnosis mechanism based on the fusion of Ho-DINA, MIRT, and IRT (LDM-HMI)}

\begin{algorithm}[!h]
	\caption{Three-channel learning diagnosis mechanism based on the fusion of Ho-DINA, MIRT, and IRT}
	\hspace*{0.02in} {\bf Input:} 
	Student exercise response matrix $R$ and question-knowledge corelation matrix $Q$\\
	\hspace*{0.02in} {\bf Output:} 
	Student exercise score prediction and interpretable parameters
	
	\begin{algorithmic}[1]
		\State 	Initialize parameters matrix EC and SC:
		\State $EC^{(0)}=\{e_{I\_diff}^{(0)},e_{I\_disc}^{(0)},e_{H\_slip}^{(0)},e_{H\_gues}^{(0)},e_{M\_disc}^{(0)},e_{M\_gues}^{(0)},e_{M\_diff}^{(0)}\} $, $SC^{(0)}=\{s_{I\_\theta}^{(0)},s_{H\_\theta}^{(0)},s_{M\_\alpha}^{(0)},s_{H\_\alpha}^{(0)}\} $
		\While{not converge or not reach iterations} 
		\State $ p(\hat{\theta}\leftarrow\theta)=min\{\frac{p(\alpha|\theta,\hat{\lambda})p(\hat{\theta})}{p(\alpha|\theta,\hat{\lambda})p(\theta)},1\} $, $ p(\hat{\alpha}\leftarrow\alpha)=min\{\frac{L(s,g;\hat{\alpha})p(\hat{\alpha}|\hat{\theta},\hat{\lambda})}{L(s,g;\alpha)p(\alpha|\hat{\theta},\hat{\lambda})},1\} $, $ p(\hat{s},\hat{g}\leftarrow s,g)=min\{\frac{L(\hat{s},\hat{g};\hat{\alpha})p(\hat{s})p(\hat{g})}{L(s,g;\hat{\alpha})p(s)p(g)},1\} $
		\State$ L_1(R)=\prod_{k=1}^{N}\prod_{j=1}^{M}p(\pi_k|\pi)p(\pi_k|\delta_j)^{r_{kj}}[1-p(\pi_k|\delta_j)]^{1-r_{kj}} $, $ L_2(R)=\prod_{i=1}^{I}\prod_{j=1}^{J}\frac{e^{r_{ij}(a_j\theta_i+d_j)}}{1+e^{r_{ij}(a_j\theta_i+d_j)}} $\\
		//update parameters in Ho-DINA
		\State$ e_{H\_slip}=\hat{s},e_{H\_gues}=\hat{g},s_{H\_\theta}=\hat{\theta},s_{H\_\alpha}=\hat{H\_\alpha}$\\
		//update parameters in IRT
		\State $e_{I\_disc}=\hat{e}_{I\_disc},e_{I\_diff}=\hat{e}_{I\_diff},s_{I\_\theta}=\hat{s}_{I\_\theta}\leftarrow \frac{\partial \log L_1(R)}{\partial e_{I\_disc}}=0,\frac{\partial \log L_1(R)}{\partial e_{I\_diff}}=0,\frac{\partial \log L_1(R)}{\partial s_{I\_\theta}}=0 $\\
		//update parameters in MIRT
		\State$ e_{M\_gues}=\hat{e}_{M\_gues},e_{M\_diff}=\hat{e}_{M\_diff},e_{M\_disc}=\hat{e}_{M\_disc},s_{M\_\alpha}=\hat{s}_{M\_\alpha}\leftarrow \frac{\partial \log L_2(R)}{\partial e_{M\_gues}}=0,\frac{\partial \log L_2(R)}{\partial e_{M\_diff}}=0,\frac{\partial \log L_2(R)}{\partial e_{M\_disc}}=0,\frac{\partial \log L_2(R)}{\partial s_{M\_\alpha}}=0 $ 
		\EndWhile
		\State$ EC^{(final)}=\{e_{I\_diff}^{(final)},e_{I\_disc}^{(final)},e_{H\_slip}^{(final)},e_{H\_gues}^{(final)},e_{M\_disc}^{(final)},e_{M\_gues}^{(final)},e_{M\_diff}^{(final)}\} $
		\State
		$ SC^{(final)}=\{s_{I\_\theta}^{(final)},s_{H\_\theta}^{(final)},s_{M\_\alpha}^{(final)},s_{H\_\alpha}^{(final)}\} $
		\For{student $s$ and exercise $e$}
		\State $h^s=f(SC_s)=\sigma(W_1 ^s\times SC_s+b_1^s)$
		\State $y^s=g(h^s)=\sigma(W_2 ^s\times h^s+b_2^s)$
		\State $h^e=f(EC_e)=\sigma(W_1 ^e\times EC_e+b_1^e)$
		\State $y^e=g(h^e)=\sigma(W_2 ^e\times h^e+b_2^e)$
		\State $h^s(\cdot),y^s(\cdot)=\mathop{\arg\min}\limits_{h^s(\cdot),y^s(\cdot)}||h^s-SC_s||^2,h^e(\cdot),y^e(\cdot)=\mathop{\arg\min}\limits_{h^e(\cdot),y^e(\cdot)}||h^e-EC_e||^2 $
		\State $ ((SC_s,h^s),(EC_e,h^e))\rightarrow(exercise,student),f=concat(exercise,student) $
		\For{$l$ in $len(f)$}
		\State $ Query = Conv(f) Key = Conv(f) Value=Conv(f) $
		\State $ f_a=\sum_{i=1}^{d_5} \frac{e^{similarity_i}}{\sum_{j=1}^{d_5}e^{similarity_j}}\cdot Value_i$
		\State $ similarity(Query,Key_i)=Query\bullet Key_i $
		\EndFor
		\EndFor
		\State $ f_p=MaxPool(Relu(Conv(f_a))) $
		\State $ p=\sigma(W_3\times f_p+b_3) $
		\State \Return $p$, $EC$, $SC$, attention weights
	\end{algorithmic}
	\label{algorithm2}
\end{algorithm}

LDM-HMI is an implementation of a three-channel learning diagnosis mechanism under our proposed intelligent learning diagnostic framework, which employs the Ho-DINA model, MIRT model, and IRT model for initial learning diagnosis. Based on the knowledge state vector of learners in the DINA model, the Ho-DINA model constructs high-order potential abilities for learners to describe learners further. Compared with the one-dimensional ability value of the IRT model, the MIRT model uses a multi-dimensional ability vector to represent learners' knowledge states. Similarly, the parameter information used to describe learners and learning resources in the three models is integrated to obtain the original input parameter set, which can be used for subsequent learning diagnosis.

\emph{\textbf{Initial diagnosis models of LDM-HMI:}} the Ho-DINA model assumes that the cognitive knowledge $\alpha_k$ is independent (partially independent) under a given ability $s_{H\_\theta}$. Cognitive knowledge has the following relationship with:
\begin{gather}
	P(\alpha \mid s_{H\_\theta})=\prod_{k=1}^{K} P(\alpha_k \mid s_{H\_\theta})\\
	P(\alpha_k \mid s_{H\_\theta})=\frac{e^{\lambda_{0k}+\lambda_{1k}s_{H\_\theta}}}{1+e^{\lambda_{0k}+\lambda_{1k}s_{H\_\theta}}}
\end{gather}
$\lambda_{k}$ represents the loading of knowledge $k$ on the ability $s_{H\_\theta}$. That is, under the DINA model, it is assumed that knowledge is locally independent and subordinate to a higher-order ability.

The response function of the MIRT model is:
\begin{equation}
	P(s_{M\_\alpha})=e_{M\_gues}+\frac{1-e_{M\_gues}}{1+e^{-D \cdot  e_{M\_{disc}}(s_{M\_\alpha}+e_{M\_diff})}}
\end{equation}
Among them, $ D $ is a constant; $ s_{M\_\alpha} $ is the learner's multi-dimensional ability vector, and each dimension is defined on a specific knowledge point; $e_{M\_{disc}}$ is the multi-dimensional discrimination vector of exercise, which is also defined in the dimension of the knowledge point; $ e_{M\_gues} $ and $ e_{M\_diff} $ are respectively guessing degree and difficulty parameters in the dimension of the question. The parameter estimation process is described in steps 1 to 18 in algorithm 2. 
The IRT model is the same as in Section 5.1. Based on this, construct a parameter set of learners and learning resources:
\begin{gather}
	EC=\{e_{I\_diff},e_{I\_disc},e_{{H\_slip}},e_{H\_gues},e_{M\_disc},e_{M\_gues},e_{M\_diff}\}\\ SC=\{s_{I\_\theta},s_{H\_\theta},s_{M\_\alpha},s_{H\_\alpha}\}
\end{gather}

\emph{\textbf{Representation and prediction network of LDM-HMI:}}
The learners and learning resources are modeled separately, and the performance prediction network is constructed to complete the learning diagnosis task in the same way as described in Section 4. The detailed steps are shown in steps 19-33 in algorithm 2.

\section{Experiment evaluation} \label{experiment}
In this section, we evaluate the learning performance prediction and interpretability attained by the proposed LDM-ID and LDM-HMI. We also provide some potential applications of the proposed method in the field of intelligent tutoring systems. The details of the experiments will be discussed below, while the codes for the experiments in this paper are released at https://github.com/CCNUZFW/LDM-ID-HMI.
\subsection{Experimental datasets}
To verify the effectiveness of the proposed intelligent learning diagnosis model, three datasets are employed to carry out the experiments. These datasets include two real-world datasets Math1\footnote{Math1:http://staff.ustc.edu.cn/\%7Eqiliuql/data/math2015.rar}, CL21\footnote{CL21: Course page: http://mooc1.chaoxing.com/course/206940184.html; Dataset released at: https://github.com/CCNUZFW/CL21} , and a virtual dataset Synthetic-5\footnote{Synthetic-5:https://github.com/chrispiech/DeepKnowledgeTracing/tree/master/data/synthetic}, of which Math1 and Synthetic-5 are both open datasets that are widely used in intelligent learning diagnosis tasks and are of moderate size, containing information such as learner-response matrices and question-knowledge point matrices that are commonly used in online intelligent learning diagnosis. CL21 is a smaller dataset of real data collected in four actual university offline classrooms and can be used to validate the applicability of the LDM framework in small offline classrooms. The details of experimental datasets are as follows:

\begin{itemize} 
	\item \textbf{Math1:} This dataset comes from several objective and subjective test questions, and answer records are collected by Wu et al. \cite{wu2015cognitive} from a college mathematics final test. For the consistency between datasets, this experiment selects the records of the learners' answers to the objective questions, which contains 4209 learners' answers on 15 exercises with 11 knowledge points.
	\item \textbf{CL21:} This dataset is collected by one of our authors, Dr. Chunyan Zeng, who is serving as the instructor of a C programming language class at the Hubei University of Technology in 2021. The dataset is collected through an intelligent tutoring system, named Chaoxing. This dataset includes the answer records of 93 learners on 36 exercises with 12 knowledge points.
	\item \textbf{Synthetic-5:} This is a virtual dataset that contains the answer records of 2,000 learners on 50 questions with 5 knowledge points.
\end{itemize}

Table \ref{tab2} shows a summary of these three datasets. The $Q$ matrices were labeled by domain experts, and the historical response matrices $R$ were collected by the intelligent tutoring systems. With the availability of the above data, the experiments were carried out with an NVIDIA GeForce GTX 2080Ti GPU in the environment of Python 3.6 and Pytorch 1.9.0.

\renewcommand{\arraystretch}{1.5} 

\begin{table}[htbp] 
	\caption{Summary of experimental datasets.}
	\label{tab2}
	\centering
	\begin{tabular}{c c c c c}
		\hline
		{\bf Dataset}&{\bf Answering records}&{\bf Number of learners}&{\bf Knowledge points}&{\bf Number of exercises}\\
		\hline
		{\bf Math1} & 63135 & 4209 &11 &15\\
		{\bf Synthetic-5}  &100000 &2000 &5 &50\\
		{\bf CL21} &3348 &93 &12 &36\\
		\hline
	\end{tabular}
\end{table}

\subsection{The methods to be compared}
Three proposals need to be evaluated: the effectiveness of learner and learning resource representation networks, the effectiveness of deep and shallow learning feature fusion, and the effectiveness of attention mechanisms. In addition, the proposed models are compared with 11 state-of-the-art methods such as item response theory (IRT) \cite{hambletonItemResponseTheory2013}, multi-dimensional item response theory (MIRT) \cite{whitely1980mirt}, deterministic input noise and door model (DINA) \cite{de2009dina}, Ho-DINA \cite{de2004hodina}, PMF-LDM \cite{Mnih2007}, neural cognitive diagnosis model (NeuralCDM) \cite{Wang2022an}, DeepMFLD \cite{ Xue2017}, and IRR \cite{Tong2021a}, which are list as follows:
\begin{itemize} 
	\item \textbf{IRT \cite{hambletonItemResponseTheory2013}}: The IRT model is a psychometrics learning diagnosis model that uses logic functions to model the relationship between the learner's learning state, the distinction, and difficulty of test questions, and the learner's response.
	
	\item \textbf{MIRT \cite{whitely1980mirt}}: The MIRT model can estimate learners' abilities in multiple dimensions at the same time, and consider the relationship between different dimensions of ability. Therefore, MIRT can estimate multi-dimensional abilities more effectively.
	
	\item \textbf{DINA \cite{de2009dina}}: The DINA model is also a psychometrics learning diagnosis model. DINA describes the learner as a multi-dimensional knowledge point mastering vector and judges the learner's answering results according to the question-knowledge correlation.
	
	\item \textbf{Ho-DINA \cite{de2004hodina}}: Based on the DINA model, the Ho-DINA model assumes that the mastery state of knowledge points is also related to the learner's general ability, which is the learner's high-level ability.
	
	\item \textbf{PMF-LDM \cite{Mnih2007}}: PMF-LDM is an intelligent learning diagnosis method based on probability matrix decomposition, which assumes that the learner's response matrix is determined by the inner product of the learner's knowledge proficiency matrix and the knowledge point vector contained in the exercise, and uses the random gradient descent method to constantly diagnose the learning performance.
	
	\item \textbf{NeuralCDM \cite{Wang2022an}}: NeuralCDM is a deep learning-based learning diagnosis method that combines neural networks to learn the complex answering process, projecting learners and exercises as latent space vectors, and using multiple neural layers to model its interaction process to predict learners answering results.
	
	\item \textbf{DeepMFLD \cite{ Xue2017}}: DeepMFLD is a matrix decomposition learning diagnosis method based on deep learning. It obtains the learners' embedding and the exercises' embedding separately by deep matrix decomposition, then fuses them with a deep neural network, and finally predicts the learner's performance by a fully connected layer.
	
	\item \textbf{IRR-IRT, IRR-MIRT, IRR-DINA, and IRR-NeuralCDM \cite{Tong2021a}}: The IRR is a general framework, which introduces pairwise learning into learning diagnosis. Four specific methods based on the IRR framework are used as the baseline: IRR-IRT, IRR-MIRT, IRR-DINA, and IRR-NCDM. The first three methods enhance the traditional psychometric diagnostic models IRT, MIRT, and DINA based on the IRR framework, while the NeuralCDM is also used to integrate with the IRR framework to form IRR-NeuralCDM.
	
\end{itemize}

\subsection{Evaluation metrics}
Since the diagnosis results of the learners' knowledge mastery state can not be judged objectively and accurately, the diagnosis results are usually used to predict the probability that the learner will answer correctly and compare it with the true value to evaluate the performance of the model. This paper uses AUC (Area Under Curve) and RMSE (Root Mean Square Error) as the evaluation indicators of the experimental results. AUC is the area under the indicator curve. The AUC value of 0.5 represents the randomly obtainable score. The higher the AUC score, the more accurate the prediction result. RMSE is used to measure the deviation between the estimated value and the true value. The smaller the value, the closer the mining result is to the true value. The calculation formula is:
\begin{equation}
	RMSE(y_i,\hat{y}_i)=\sqrt{\frac{1}{m}\sum_{i=1}^{m}{(y_i-\hat{y}_i)}^2 } 
\end{equation}
Among them, $m$ represents the size of the test data set, $ y_i $ is the true value of the student's answer, and $\hat{y}_i$ is the predicted value of the student's answer. 

\subsection{Experimental results and analysis}
In our experiments, a 5-fold cross-validation is performed, and the average AUC and RMSE of five times experiments are used as the evaluation metrics. Furthermore, 80$\%$ samples of each dataset are employed as the training set, and 20$\%$ of the data are left for the test to check whether the proposed method is generalized well.

\subsubsection{Comparision with the state-of-the-art methods.}

\begin{table}[!h]
	\centering
	\caption{Comparison with the state-of-the-art methods in terms of learning performance prediction.}
	\begin{tabular}{lcccccc}
		\toprule
		\multirow{2}[4]{*}{\textbf{Model}} & \multicolumn{2}{c}{\textbf{CL21}} & \multicolumn{2}{c}{\textbf{Math1}} & \multicolumn{2}{c}{\textbf{Synthetic-5}} \\
		\cmidrule{2-7}          & \textbf{AUC} & \textbf{RMSE} & \textbf{AUC} & \textbf{RMSE} & \textbf{AUC} & \textbf{RMSE} \\
		\midrule
		IRT \cite{hambletonItemResponseTheory2013}   & 73.12\% & 0.4365  & 68.31\% & 0.5164  & 78.52\% & 0.4824  \\
		MIRT \cite{whitely1980mirt}  & 73.24\% & 0.4352  & 69.83\% & 0.4946  & 79.80\% & 0.4733  \\
		DINA \cite{de2009dina} & 69.92\% & 0.4859  & 64.46\% & 0.5297  & 73.80\% & 0.4968  \\
		Ho-DINA \cite{de2004hodina} & 70.37\% & 0.4723  & 66.58\% & 0.5215  & 72.45\% & 0.4988  \\
		PMF-LDM \cite{Mnih2007} & 63.74\% & 0.6515  & 68.03\% & 0.4884  & 62.40\% & 0.6038  \\
		NeuralCDM \cite{Wang2022an} & 76.88\% & 0.4274  & 72.90\% & 0.4533  & 79.77\% & 0.4756  \\
		DeepMFLD \cite{ Xue2017} & 80.40\% & 0.4226  & 77.74\% & 0.4356  & 76.68\% & 0.4897  \\
		IRR-IRT \cite{Tong2021a} & 71.23\% & 0.4395  & 74.10\% & 0.4505  & 72.65\% & 0.4996 \\
		IRR-MIRT \cite{Tong2021a} & 81.29\% & 0.4213 & 80.72\% & 0.4231 & 81.35\% & 0.4073 \\
		IRR-DINA \cite{Tong2021a} & 75.51\% & 0.4271 & 81.40\% & 0.4234 & 81.03\% & 0.4087 \\
		IRR-NeuralCDM \cite{Tong2021a} & 75.02\% & 0.4279 & 78.56\% & 0.4278 & 75.85\% & 0.4917 \\
		Proposed LDM-ID & \textbf{86.44\%} & \textbf{0.4105 } & 81.35\% & 0.4171 & 88.02\% & 0.3667 \\
		Proposed LDM-HMI & 85.91\% & 0.4122  & \textbf{81.89\%} & \textbf{0.4163 } & \textbf{90.25\%} & \textbf{0.3528} \\
		\bottomrule
	\end{tabular}%
	\label{tab3}%
\end{table}%

For the LDM-ID and LDM-HMI methods, detailed settings are as follows: Sigmoid is used as the activation function of the fully connected layer, and the CNN in the prediction network consists of a convolutional layer, a Relu activation layer, and a maximum pooling layer, where the convolutional kernel size is 3. The model is parameterized by the Adam optimizer, the learning rate is set to 0.001, and the dropout ratio is set to 0.2. In the learner representation network and learning resource representation network, the learner potential vector and learning resource potential vector dimensions are set to 128 and 64, respectively. In addition, in the baseline models, 3PL-IRT is used for comparison, and the capability latitude value is set to 3 in the MIRT model.

As shown in Table \ref{tab3}, in terms of the accuracy of learning performance prediction, among the traditional psychometric-based learning diagnostic models, MIRT outperformed IRT, DINA, Ho-DINA, and PMF-LDM overall on the three datasets. The performance of NeuralCDM, DeepMFLD, IRR and the proposed methods LDM-ID, LDM-HMI generally outperformed the psychometric-based learning diagnostic models, which indicates that the deep network-based learning diagnostic methods outperform the traditional psychometric-based learning diagnostic methods. Usually, deep networks have more complex structures and more parameters, and the information they can characterize will be richer, while psychometric-based methods usually have simpler structures, fewer parameters, and limited characterization capabilities.



\begin{figure}[htbp]
	\centering
	\includegraphics[width=\linewidth]{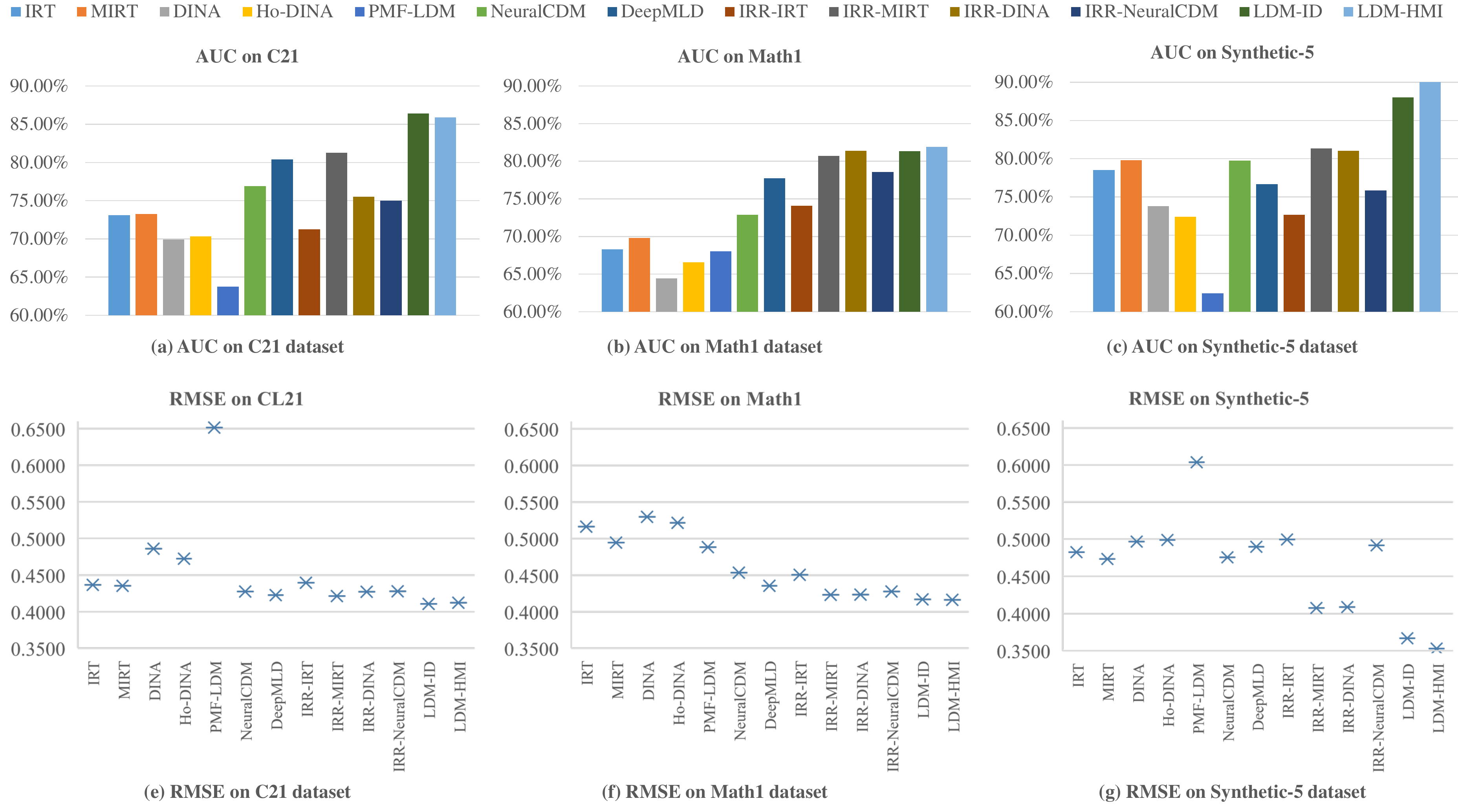}
	\caption{The proposed methods were compared with 11 state-of-the-art methods. In this paper, 11 state-of-the-art learning diagnosis methods were replicated and experiments were conducted on two publicly available datasets and one self-built dataset, and evaluated using two metrics, AUC and RMSE. A larger AUC indicates higher prediction accuracy, and a smaller RMSE indicates better stability of the methods.}
	\label{fig2}
\end{figure}

As shown in Figure \ref{fig2}, the LDM-ID and LDM-HMI methods proposed in this paper outperformed all baseline models in terms of the accuracy of learning performance prediction. On the three datasets, their AUC improved by about 11.45\% and RMSE decreased by about 0.07 compared to MIRT on average, their AUC improved by about 8.57\% and RMSE decreased by about 0.06 compared to NeuralCDM on average. On the one hand, the cognitive parameters in this paper provide more effective input information for the diagnosis and prediction tasks from the source. On the other hand, this paper constructs a more capable network of learners and learning resources and explores the potential characteristics of learners and learning resources more effectively.

Further, among the two implementations of the LDM framework proposed in this paper, LDM-HMI performs better on the larger-scale datasets Math1 and Synthetic-5, with an AUC about 1.39\% higher and an RMSE about 0.007 lower than that of LDM-ID over the two datasets, while on the smaller-scale dataset CL21, LDM-ID performs better, with an AUC about 0.53\% higher and an RMSE about 0.002 lower than that of LDM-HMI. This indicates that the LDM-ID method may be more suitable for small data volume scenarios. In other words, under the LDM framework of this paper, the specific implementation can be flexibly adjusted according to the requirements to achieve better adaptation to the actual scenarios.

As shown in Table \ref{tab4}, the LDM-ID and LDM-HMI proposed in this paper are comparable to the deep learning diagnostic method NeuralCDM, DeepMFLD, and IRR in terms of time loss. The prediction time loss of the proposed method on the medium-sized dataset Math1 is comparable to that of the traditional psychometric methods. On the small dataset CL21, and large dataset Synthetic-5, the time loss of the proposed method is one order of magnitude higher compared to the traditional psychometric methods. Overall, the prediction time of the proposed method is about 0.2 ms on a single sample, which is sufficient for practical educational application scenarios.


\begin{table}[htbp]
	\centering
	\caption{Comparison with traditional methods in terms of prediction time cost.}
	\begin{tabular}{lccc}
		\toprule
		\multirow{2}[4]{*}{\textbf{Model}} & \multicolumn{3}{c}{\textbf{Time cost for prediction}} \\
		\cmidrule{2-4}          & \textbf{CL21} & \textbf{Math1} & \textbf{Synthetic-5} \\
		\midrule
		IRT \cite{hambletonItemResponseTheory2013}  & 16 ms  & 203 ms & 141 ms \\
		MIRT \cite{whitely1980mirt} & 90 ms  & 385 ms & 597 ms \\
		DINA \cite{de2009dina} & 16 ms  & 31 ms  & 47 ms \\
		Ho-DINA \cite{de2004hodina} & 57 ms  & 198 ms & 124 ms \\
		PMF-LDM \cite{Mnih2007} & 15 ms & 45 ms & 61 ms \\
		NeuralCDM \cite{Wang2022an} & 156 ms & 216 ms & 3359 ms \\
		DeepMFLD \cite{ Xue2017} & 270 ms & 352 ms & 3822 ms \\
		IRR-IRT \cite{Tong2021a} & 192 ms & 401 ms & 3591 ms \\
		IRR-MIRT \cite{Tong2021a} & 229 ms & 435 ms & 3603 ms \\
		IRR-DINA \cite{Tong2021a} & 223 ms & 462 ms & 3609 ms \\
		IRR-NeuralCDM \cite{Tong2021a} & 218 ms & 316 ms & 3583 ms \\
		Proposed LDM-ID & 187 ms & 277 ms & 3397 ms \\
		Proposed LDM-HMI & 203 ms & 303 ms & 3374 ms \\
		\bottomrule
	\end{tabular}%
	\label{tab4}%
\end{table}%

\subsubsection{Effectiveness of Learner and Learning Resource Representation Networks}
To verify the effectiveness of the learner and learning resource characterization network in the interpretable intelligent learning diagnosis framework proposed in this paper, experiments are conducted to compare the auto-encoder-based characterization network in this paper with the fully connected deep neural network in the traditional approach.

As shown in Table \ref{tab5}, in the LDM-ID and LDM-HMI methods, compared with the fully connected deep neural network, the auto-encoder-based representational network improves the AUC of the model by about 0.85\% and 0.9\% on average over the three data sets, respectively, and reduces the RMSE by about 0.002 and 0.003, respectively. It proves that the auto-encoder-based learner and learning resource representation network is more effective. Among them, the encoder learns to retain as much relevant information in the potential space and discard irrelevant parts, and the decoder learns the potential space information to reconstruct the input, which achieves an effective denoising effect and contributes to a more effective potential feature representation of learners and learning resources.

\begin{table}[htbp]
	\centering
	\caption{Learning performance prediction experiments for learner and learning resource representation networks.}
	\begin{tabular}{ccccccccc}
		\toprule
		\multirow{2}[4]{*}{\textbf{Model}} & \multicolumn{2}{c}{\multirow{2}[4]{*}{\textbf{Network Structure}}} & \multicolumn{2}{c}{\textbf{CL21}} & \multicolumn{2}{c}{\textbf{Math 1}} & \multicolumn{2}{c}{\textbf{Synthetic-5}} \\
		\cmidrule{4-9}          & \multicolumn{2}{c}{} & \textbf{AUC} & \textbf{RMSE} & \textbf{AUC} & \textbf{RMSE} & \textbf{AUC} & \textbf{RMSE} \\
		\midrule
		\multirow{2}[2]{*}{\textbf{LDM-ID}} & \multicolumn{2}{c}{DNN} & 84.78\% & 0.4179 & 79.26\% & 0.4223 & 86.15\% & 0.3691 \\
		& \multicolumn{2}{c}{SAE} & \textbf{85.32\%} & \textbf{0.4157} & \textbf{80.47\%} & \textbf{0.4184} & \textbf{86.96\%} & \textbf{0.3682} \\
		\midrule
		\multirow{2}[2]{*}{\textbf{LDM-HMI}} & \multicolumn{2}{c}{DNN} & 81.56\% & 0.4219 & 79.74\% & 0.4215 & 88.60\% & 0.3579 \\
		& \multicolumn{2}{c}{SAE} & \textbf{82.65\%} & \textbf{0.4201} & \textbf{81.02\%} & \textbf{0.4168} & \textbf{88.93\%} & \textbf{0.3562} \\
		\bottomrule
	\end{tabular}%
	\label{tab5}%
\end{table}%

\subsubsection{Effectiveness of Deep and Shallow Learning Feature Fusion}
To verify the effectiveness of the fusion of deep and shallow features in the interpretable intelligent learning diagnosis framework proposed in this paper, experiments are conducted in three forms: using only shallow features, using only deep features, and fusing deep and shallow features.

As shown in Table \ref{tab6}, the fusion of deep and shallow features has the most significant effect on the model effectiveness on the average over the three datasets, with the AUC of the fusion of deep and shallow features improving by about 1.46\% and 1.73\% and the RMSE decreasing by about 0.011 and 0.016 in the LDM-ID and LDM-HMI approaches relative to the use of only shallow features, relative to the use of only deep features, the AUC of deep and shallow feature fusion is increased by about 0.42\% and 0.72\%, and the RMSE is reduced by about 0.003 and 0.011. Therefore, the fusion of deep and shallow features helps to improve the performance of the model, because the deep and shallow features provide different valid information from the high-dimensional space and low-dimensional space, respectively, to achieve better prediction results. Also, the fusion of deep and shallow features has better interpretation compared to using only deep features.

\begin{table}[htbp]
	\centering
	\caption{Learning performance prediction experiments with a fusion of deep and shallow features.}
	\begin{tabular}{ccccccccc}
		\toprule
		\multirow{2}[4]{*}{\textbf{Model}} & \multicolumn{2}{c}{\multirow{2}[4]{*}{\textbf{Representation}}} & \multicolumn{2}{c}{\textbf{CL21}} & \multicolumn{2}{c}{\textbf{Math 1}} & \multicolumn{2}{c}{\textbf{Synthetic-5}} \\
		\cmidrule{4-9}          & \multicolumn{2}{c}{} & \textbf{AUC} & \textbf{RMSE} & \textbf{AUC} & \textbf{RMSE} & \textbf{AUC} & \textbf{RMSE} \\
		\midrule
		\multirow{3}[2]{*}{\textbf{LDM-ID}} & \multicolumn{2}{c}{shallow feature} & 83.16\% & 0.4291 & 77.41\% & 0.4388 & 85.23\% & 0.3749 \\
		& \multicolumn{2}{c}{deep feature} & 84.31\% & 0.4206 & 78.69\% & 0.4265 & 85.91\% & 0.3718 \\
		& \multicolumn{2}{c}{shallow + deep feature} & \textbf{84.78\%} & \textbf{0.4179} & \textbf{79.26\%} & \textbf{0.4223} & \textbf{86.15\%} & \textbf{0.3691} \\
		\midrule
		\multirow{3}[2]{*}{\textbf{LDM-HMI}} & \multicolumn{2}{c}{shallow feature} & 79.66\% & 0.4435 & 78.33\% & 0.4397 & 86.73\% & 0.3664 \\
		& \multicolumn{2}{c}{deep feature} & 81.04\% & 0.4359 & 79.21\% & 0.4342 & 87.48\% & 0.3638 \\
		& \multicolumn{2}{c}{shallow + deep feature} & \textbf{81.56\%} & \textbf{0.4219} & \textbf{79.74\%} & \textbf{0.4215} & \textbf{88.60\%} & \textbf{0.3579} \\
		\bottomrule
	\end{tabular}%
	\label{tab6}%
\end{table}%

\subsubsection{Effectiveness of attention mechanism}
Based on the deep and shallow feature fusion, considering that each dimension of deep features and shallow features may have different effects on the performance of model prediction, giving different degrees of attention to different dimensional features may produce better results. Therefore, this paper introduces an attention mechanism in the prediction network to further process the deep and shallow fused features, and conducts experimental validation.

As shown in Table \ref{tab7}, after introducing the attention mechanism, the AUC is improved by about 0.92\% and 1.42\% and the RMSE is reduced by about 0.004 and 0.005 on average in the three datasets in the LDM-ID and LDM-HMI methods, which fully illustrates that the introduction of the attention mechanism can improve the performance of the model. The attention mechanism assigns different weights to the features of different dimensions, which not only helps to improve the accuracy of the prediction results but also can provide explanatory information for the prediction results. It can help us to know the potential connection between the features and the output.

\begin{table}[htbp]
	\centering
	\caption{Experiments on the prediction of learning performance of attention mechanism.}
	\begin{tabular}{ccccccccc}
		\toprule
		\multirow{2}[4]{*}{\textbf{Model}} & \multicolumn{2}{c}{\multirow{2}[4]{*}{\textbf{Attention Setting}}} & \multicolumn{2}{c}{\textbf{CL21}} & \multicolumn{2}{c}{\textbf{Math 1}} & \multicolumn{2}{c}{\textbf{Synthetic-5}} \\
		\cmidrule{4-9}          & \multicolumn{2}{c}{} & \textbf{AUC} & \textbf{RMSE} & \textbf{AUC} & \textbf{RMSE} & \textbf{AUC} & \textbf{RMSE} \\
		\midrule
		\multirow{2}[2]{*}{\textbf{LDM-ID}} & \multicolumn{2}{c}{without attention} & 84.78\% & 0.4179 & 79.26\% & 0.4223 & 86.15\% & 0.3691 \\
		& \multicolumn{2}{c}{with attention} & \textbf{85.39\%} & \textbf{0.4123} & \textbf{79.84\%} & \textbf{0.4192} & \textbf{87.71\%} & \textbf{0.3669} \\
		\midrule
		\multirow{2}[2]{*}{\textbf{LDM-HMI}} & \multicolumn{2}{c}{without attention} & 81.56\% & 0.4219 & 79.74\% & 0.4215 & 88.60\% & 0.3579 \\
		& \multicolumn{2}{c}{with attention} & \textbf{84.70\%} & \textbf{0.4134} & \textbf{80.33\%} & \textbf{0.4185} & \textbf{89.14\%} & \textbf{0.3537} \\
		\bottomrule
	\end{tabular}%
	\label{tab7}%
\end{table}%

In addition, this paper also verifies the effect of the joint action of the auto-encoder-based learner and learning resource representation network and attention mechanism based on deep and shallow feature fusion, which is conducted in three cases, namely, introducing the auto-encoder-based learner and learning resource representation network alone, introducing the attention mechanism alone, and introducing the auto-encoder-based learner and learning resource representation network and attention mechanism simultaneously. The experiments were conducted.

\begin{table}[!h]
	\centering
	\caption{Learning performance prediction experiments for the representation network and attention mechanism.}
	\begin{tabular}{ccccccccc}
		\toprule
		\multirow{2}[4]{*}{\textbf{Model}} & \multicolumn{2}{c}{\multirow{2}[4]{*}{\textbf{Attention Setting}}} & \multicolumn{2}{c}{\textbf{CL21}} & \multicolumn{2}{c}{\textbf{Math 1}} & \multicolumn{2}{c}{\textbf{Synthetic-5}} \\
		\cmidrule{4-9}          & \multicolumn{2}{c}{} & \textbf{AUC} & \textbf{RMSE} & \textbf{AUC} & \textbf{RMSE} & \textbf{AUC} & \textbf{RMSE} \\
		\midrule
		\multirow{3}[2]{*}{\textbf{LDM-ID}} & \multicolumn{2}{c}{SAE} & 85.32\% & 0.4157 & 80.47\% & 0.4184 & 86.96\% & 0.3682 \\
		& \multicolumn{2}{c}{ attention} & 85.39\% & 0.4123 & 79.84\% & 0.4192 & 87.71\% & 0.3669 \\
		& \multicolumn{2}{c}{SAE with attention} & \textbf{86.44\%} & \textbf{0.4105} & \textbf{81.35\%} & \textbf{0.4171} & \textbf{88.02\%} & \textbf{0.3667} \\
		\midrule
		\multirow{3}[2]{*}{\textbf{LDM-HMI}} & \multicolumn{2}{c}{SAE} & 82.65\% & 0.4201 & 81.02\% & 0.4168 & 88.93\% & 0.3562 \\
		& \multicolumn{2}{c}{ attention} & 84.70\% & 0.4134 & 80.33\% & 0.4185 & 89.14\% & 0.3537 \\
		& \multicolumn{2}{c}{SAE with attention} & \textbf{85.91\%} & \textbf{0.4122} & \textbf{81.89\%} & \textbf{0.4163} & \textbf{90.25\%} & \textbf{0.3528} \\
		\bottomrule
	\end{tabular}%
	\label{tab8}%
\end{table}%

As shown in Table \ref{tab8}, the model has the best performance when the auto-encoder-based learner and learning resource representation network and the attention mechanism are introduced simultaneously. The AUC improves by 1.02\% and 1.82\%, and the RMSE decreases by 0.003 and 0.004 in the LDM-ID and LDM-HMI methods on average of the three datasets compared to the auto-encoder-based learner and learning resource representation network alone. Compared with the attention mechanism alone, the AUC of the model increased by 0.96\% and 1.29\%, and the RMSE decreased by 0.001 and 0.001. This indicates that the model performs best when the auto-encoder-based learner and learning resource representation network and the attention mechanism are combined, which can effectively represent the learners and learning resources while assigning different weights to the features of different dimensions.

\subsubsection{Interpretable analysis and educational applications}
The interpretable intelligent learning diagnostic framework proposed in this paper has good interpretability, which comes from three main aspects: cognitive parameters, learner-resource response network, and weighting information of attention mechanisms.

\textit{1) Interpretability of cognitive parameters}

In the initial part of the model, this paper conducts a preliminary diagnostic evaluation of learners based on different cognitive diagnostic model, and obtains some cognitive parameters that have practical significance in the educational field, such as the multi-dimensional ability values of learners in the MIRT model, the binary vector of learners' knowledge mastery in the DINA model, the difficulty, differentiation, the skipping rate and guessing rate of learning resources. These parameters can complement each other and form a comprehensive and multi-dimensional description of the learners and learning resources, which can be used as input to the subsequent prediction network to provide a better explanation.

\begin{figure}[!h]
	\centering
	\includegraphics[width=\linewidth]{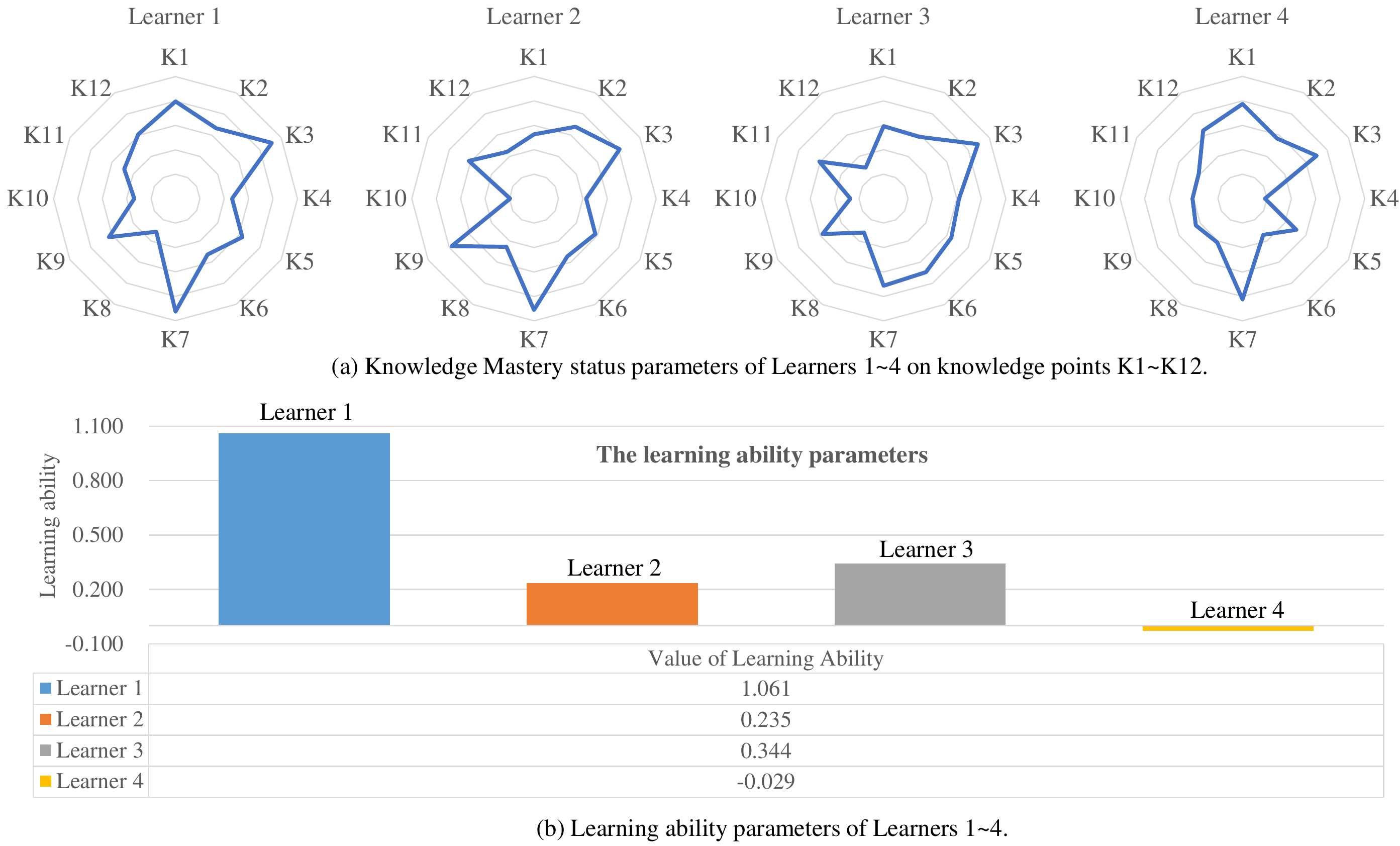}
	\caption{There are two sets of interpretable cognitive parameters for learners, the knowledge mastery cognitive parameters and the learning ability parameters. For the knowledge mastery cognitive parameter, 0 represents no mastery of the knowledge, and 1 represents full mastery of the knowledge. For the cognitive parameter of learning ability, the range of values is from -1.5 to 1.5, representing low to high learning ability. The higher the learning ability parameters, the higher the probability of answering the exercise correctly.}
	\label{fig3}
\end{figure}

\begin{figure*}
	\centering
	\subfigure[Cognitive parameters of 20 exercises.]{
		\begin{minipage}{0.345\linewidth}
			\centering
			\includegraphics[width=\linewidth]{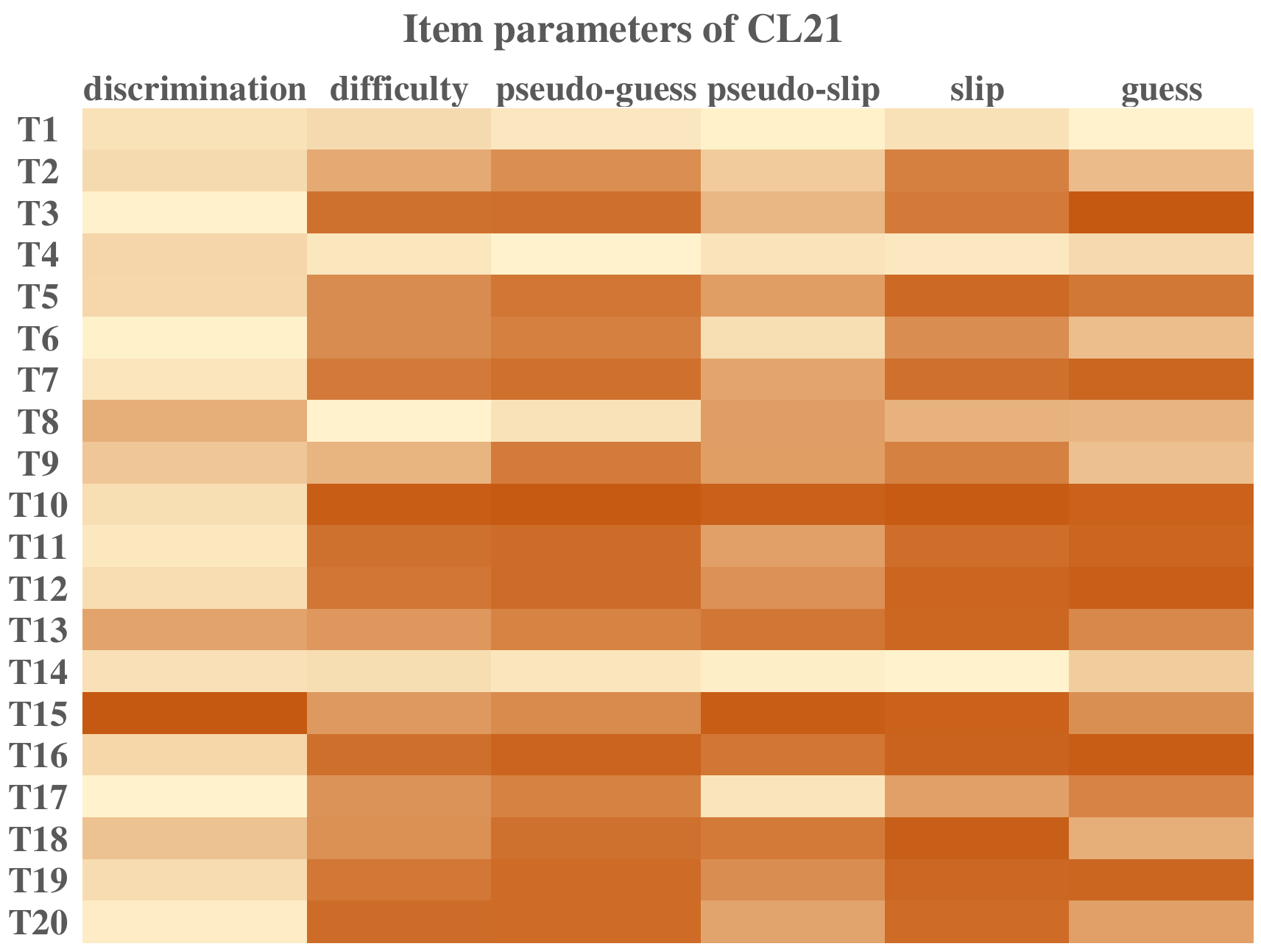}
			\label{fig4a}
		\end{minipage}
	}
	\subfigure[Heatmap of learner-resource response network.]{
		\begin{minipage}{0.31\linewidth}
			\centering
			\includegraphics[width=\linewidth]{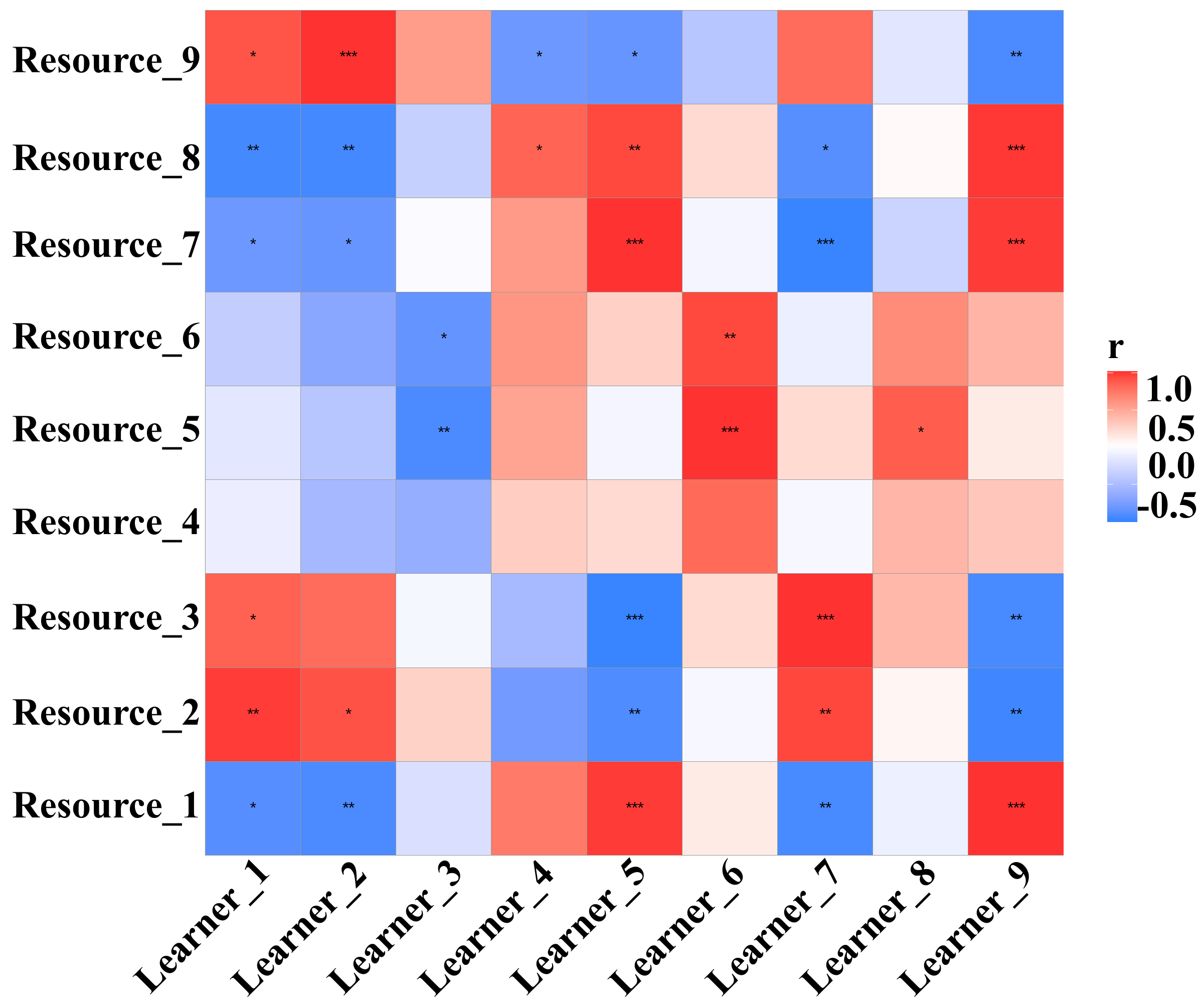}
			\label{fig4b}
		\end{minipage}
	}
	\subfigure[Heatmap of attention weighting factors.]{
		\begin{minipage}{0.305\linewidth}
			\centering
			\includegraphics[width=\linewidth]{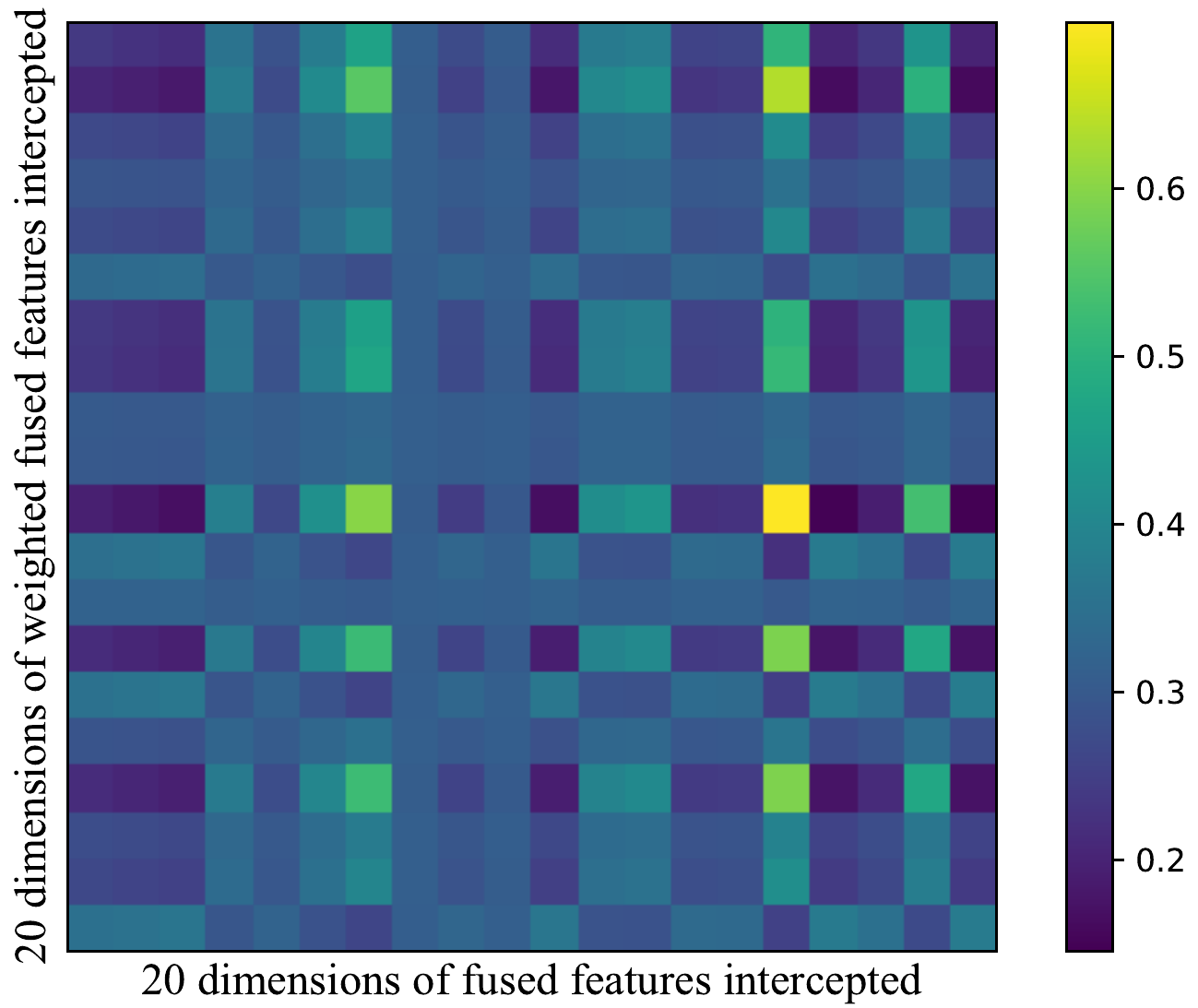}
			\label{fig4c}
		\end{minipage}
	}
	\caption{As in Figure \ref{fig4a}, 6 groups of the relevant cognitive parameters of 20 exercises are intercepted from CL21. The darker the color is, the greater the value is. For Figure \ref{fig4b}, the correlation coefficients between learner embedding and learning resource embedding are extracted in the learner-resource response network. For Figure \ref{fig4c}, the weight factors of the self-attention mechanism on the learning feature vector are used to speculate the answering results.}
\end{figure*}

For example, for learners, the proposed interpretable intelligent learning diagnostic framework can be combined with multi-perspective cognitive diagnostic theories to evaluate learners' learning states, which can not only understand the differences in learners' multidimensional and fine-grained potential cognition status but also obtain the overall ability level of learners. In addition, it can provide refined tutoring information for subsequent teaching practice. Figure \ref{fig3} (a) shows the knowledge mastery of four randomly selected learners in the CL21 dataset and their overall proficiency level. It shows that learner $S1$ has a low mastery of knowledge points $K8$ and $K10$. Figure \ref{fig3} (b) shows the overall proficiency level of the four learners, and we can see that the overall proficiency level of learner $S4$ is low, indicating that his foundation may be weak and he needs to consolidate his basic knowledge. Figure \ref{fig4a} visualizes six cognitive parameters of 20 exercises on the CL21 dataset. This information can be effectively applied in various aspects, such as learner profiling and learner proficiency analysis.

Secondly, for learning resources, this method can provide comprehensive parameter characterization information of learning resources. Based on these parameters, a more accurate learning resource recommendation service can be provided for learners. For example, when learners have a higher mastery of knowledge point $a$, we can recommend more difficult learning resources to help learners master knowledge point a more firmly, or when there is a small difference in the diagnostic results of different learners' mastery of knowledge point $b$, we can select more differentiated learning resources about $b$ to provide detailed learning diagnosis for these learners. This rich information about learning resources can assist in practical applications such as educational resource modeling and test question quality analysis.


\textit{2) Interpretability of learner-resource response network} 

After deep characterization of learners and learning resources, this paper constructs a learner-learning resource response network to model the learner-learning resource response process in the actual scenario, simulating the interrelationship between the potential learning state of learners and the response results in the response process, which is more suitable for the actual situation and provides explanations in the modeling method. The heat map of the Pearson coefficient correlation between the learner characteristics and the learning resource characteristics is as follows, and Figure \ref{fig4b} shows that there is a more significant connection between the two. The learner-learning resource response network can be applied to learning resource recommendation, learning interaction analysis, and adaptive learning systems.


\textit{3) Interpretability of attention mechanisms} 

Based on the model's deep and shallow integration features of learners and learning resources, an attention mechanism is introduced to obtain the difference in the degree of influence of features of different dimensions on the actual response results of learners, so that the importance level of features provides retrospective explanatory information for predicting results, and provides support for educational cue analysis, educational visualization research, etc.

Secondly, from the perspective of learning resources, this method can provide comprehensive parameter characterization information for learning resources. Figure \ref{fig4c} shows the exercise parameter information of the CL21 data set. Based on these exercise parameters, it is possible to provide learners with more accurate learning resource recommendation services. For example, when the learner has a high degree of mastery of knowledge point a, we can recommend difficult topic resources about knowledge point a to help learners grasp this knowledge point more firmly, or when different learners have a small difference in the diagnosis result of the mastery of knowledge point b, we can make a more detailed learning diagnosis for these learners by selecting exercise resources with a greater degree of discrimination.

\section{Conclusion} \label{conclusion}
To address the challenge of balancing diagnostic accuracy and interpretability in traditional learning diagnostic methods, this paper proposes a unified interpretable intelligent learning diagnosis framework that integrates the interpretability of cognitive diagnosis methods and the powerful representation learning ability of deep learning methods, which uses learners' learning records and learning resources related information to perform learning diagnosis and learning performance prediction for learners. Using the diagnostic results of multi-channel prior cognitive diagnosis as the basis for embedding representation, a learner-learning resource response network is constructed to simulate the process of learners' answering behaviors. More important influencing factors for prediction results are mined through the self-attention mechanism, making the model more interpretable.

Specifically, in the modeling process, we first construct cognitive parameter sets to form shallow features based on educational theories from multiple cognitive diagnosis models. Then design a learner representation network and learning resource representation network to mine the deep feature of learners and learning resources, to facilitate the use of shallow features in combination with deep features. In the diagnosis process, a fusion self-attention mechanism based on convolutional neural network architecture is used for learners' performance prediction, which is helpful in finding the features that have a greater impact on the learner's performance. Experiments on real-world data sets verify the effectiveness of the proposed method, and the analysis of the experimental results can prove to a certain extent the possibility of the method proposed in this paper in real education scenarios.

Our method can obtain the predicted value of the correct probability of the learner's answer, diagnose the learner's overall knowledge point mastery through the characteristic information of the learner and the learning resource, conduct the parameter representation of the learning resource, and provide reference information for the individualized learning of the learner and the intelligent recommendation of the learning resource of the learning platform. 
Future research can be carried out from the following aspects: \textit{1)} In the intelligent learning diagnosis framework of this paper, more learning diagnosis models could be introduced, and different methods and perspectives could be used to model learners and learning resources; \textit{2)} It could monitor the learner's learning process more macroscopically and evaluate the learner's learning in periods, and dynamically observe the change process of the learner's knowledge mastery status and learning ability during learning; \textit{3)} It could introduce multi-modal data, such as the text, video, and audio information in the learning resources. The learner's behavior information could be used to enrich the input data of the model to obtain more accurate learning evaluation results.

\section*{acknowledgements}
This work is supported by the National Natural Science Foundation of China (No. 62177022, 61901165), AI and Faculty Empowerment Pilot Project (No. CCNUAI\&FE2022-03-01), Collaborative Innovation Center for Informatization and Balanced Development of K-12 Education by MOE and Hubei Province (No. xtzd2021-005), and National Natural Science Foundation of China (No. 61501199).

\section*{conflict of interest}
No potential conflict of interest was reported by the authors.

\section*{endnotes}
Math1: \url{http://staff.ustc.edu.cn/\%7Eqiliuql/data/math2015.rar}
\\
Synthetic-5: \url{https://github.com/chrispiech/DeepKnowledgeTracing/tree/master/data/synthetic}
\\
CL21: \url{https://github.com/CCNUZFW/CL21}
\\
Codes for LDM-ID and LDM-HMI: \url{https://github.com/CCNUZFW/LDM-ID-HMI}

\bibliographystyle{WileyNJD-AMA}
\bibliography{refs,ALLBib}

\end{document}